%% file: main.tex
\documentclass{IEEEtran}

\usepackage[utf8]{inputenc} % allow utf-8 input
\usepackage[T1]{fontenc}    % use 8-bit T1 fonts
\usepackage{hyperref}       % hyperlinks
\usepackage{url}            % simple URL typesetting
\usepackage{rotating}
\usepackage{booktabs}       % professional-quality tables
\usepackage{amsfonts}       % blackboard math symbols
\usepackage{nicefrac}       % compact symbols for 1/2, etc.
\usepackage{microtype}      % microtypography
\usepackage{subfiles}
\usepackage{copyrightbox}
\usepackage[export]{adjustbox}
\usepackage{ragged2e}
\usepackage{floatrow}
\usepackage{graphicx, subcaption}
\usepackage{amsmath}
\usepackage{wrapfig}
\usepackage{multirow}
\usepackage{floatrow}
\usepackage{caption}
\usepackage{booktabs}
\usepackage{subcaption}
\usepackage{amssymb}
\DeclareMathOperator*{\argmax}{arg\,max}
\DeclareMathOperator*{\argmin}{arg\,min}
\title{Unsupervised Image-to-Image Translation via Pre-trained StyleGAN2 Network}

\author{%
    Jialu Huang, Jing Liao, Sam Kwong, Fellow, IEEE\\
}

\begin{document}
\maketitle
%\twocolumn[{%
%\renewcommand\twocolumn[1][]{#1}%
%\maketitle
%\begin{center}
%    \centering
%    \includegraphics[width=0.9\textwidth]{fig/T.pdf}
%    \captionof{figure}{\small{We propose a multi-density sketch-to-image translation network (MDSIT), which can support input sketches at different density levels (Top). Our MDSIT can be applied in applications including face editing and anime colorization, and also provide coarse-to-fine levels of controls to these applications (Bottom).}}
%\end{center}%
%}]

\begin{abstract}
Image-to-Image (I2I) translation is a heated topic in academia, and it also has been applied in real-world industry for tasks like image synthesis, super-resolution, and colorization. However, traditional I2I translation methods train data in two or more domains together. This requires lots of computation resources. Moreover, the results are of lower quality, and they contain many more artifacts. The training process could be unstable when the data in different domains are not balanced, and modal collapse is more likely to happen. We proposed a new I2I translation method that generates a new model in the target domain via a series of model transformations on a pre-trained StyleGAN2 model in the source domain. After that, we proposed an inversion method to achieve the conversion between an image and its latent vector. By feeding the latent vector into the generated model, we can perform I2I translation between the source domain and target domain.  Both qualitative and quantitative evaluations were conducted to prove that the proposed method can achieve outstanding performance in terms of image quality, diversity and semantic similarity to the input and reference images compared to state-of-the-art works.
\end{abstract}

\let\thefootnote\relax\footnotetext{This work was supported in part by the National Natural Science Foundation of China Grant 61672443, in part by Hong Kong GRF-RGC General Research Fund under Grant 9042322 (CityU 11200116), Grant 9042489 (CityU 11206317), and Grant 9042816 (CityU 11209819)\par
This work was supported in part by the Hong Kong Research Grants Council (RGC) Early Career Scheme under Grant 9048148 (CityU 21209119), and in part by the CityU of Hong Kong under APRC Grant 9610488.\par
Jialu Huang, Jingliao and Sam Kwong are with the Department of Computer Science, City University of Hong Kong, Kowloon, Hong Kong (e-mail: jialhuang8-c@my.cityu.edu.hk, jingliao@cityu.edu.hk, cssamk@cityu.edu.hk) Sam Kwong is also with City University of Hong Kong Shenzhen Research Institute.\par Implementation can be found at: \textcolor{blue}{\url{https://github.com/HideUnderBush/UI2I_via_StyleGAN2}}}

\newcommand{\Lim}[1]{\raisebox{0.5ex}{\scalebox{0.8}{$\displaystyle \lim_{#1}\;$}}}
\section{Introduction}
\input{sections/introduction.tex}
\section{Related Work} \label{sec:rela}
\input{sections/related_work.tex}
\section{Method}
\input{sections/method.tex}
\section{Experiment}
\input{sections/experiments.tex}
%\section{Application}
%\input{sections/applications.tex}
\section{Conclusion}
We proposed an unsupervised I2I translation method via pre-trained StyleGAN2 network, which can generate images in target domains with high quality. We first defined the model distance and tried to generate a model from a base StyleGAN2 model in the target domain through a series of model transformations. The transformed model has a relatively small distance from the base model. Then we proposed an inversion method to obtain the latent vector from an image, based on a specific StyleGAN2 model. This was then injected to the transformed StyleGAN2 network to achieve I2I translation. Our method supports multi-modal and multi-domain I2I translation with less computation resources including training time and memory. In addition, it achieves much higher performance in terms of image fidelity, diversity and semantic similarity to the input or reference images than traditional I2I translation methods. This is because the proposed method focuses on one domain at a time, while traditional I2I methods train models with data in various domains, which may cause lower resolution, artifacts and modal collapse. Moreover, both qualitative and quantitative results showed that the proposed method has obvious superiority than state-of-the-art work. Although the proposed method can generate outstanding results, it does require relatively long time in test stage as the optimization based inversion method cost relatively long time then feed-forward methods. We aim to improve the performance in future work.  \par

% Bibliography
\bibliography{reference}

\end{document}

%% file: sections/introduction.tex
Image-to-Image (I2I) translation has been a hot topic recently. With rapid development of deep learning, I2I translation has been greatly improved in the past few years. It also provides solutions to real-world applications like de-noise, de-rain, colorization, super-resolution and image editing.\par
% figure example 
\begin{figure*}[!ht]
   \centering
%---------row 1
\begin{subfigure}{4.2cm}
\includegraphics[width=4.2cm]{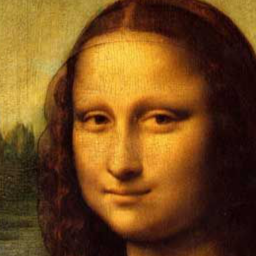}\caption*{\small{(a) input portrait}}
\end{subfigure}
\begin{subfigure}{4.2cm}
\includegraphics[width=4.2cm]{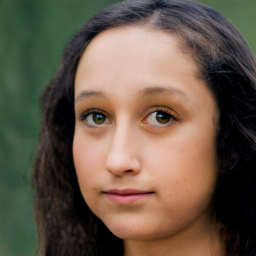}\caption*{\small{(b) portrait2face}}
\end{subfigure}
\begin{subfigure}{4.2cm}
\includegraphics[width=4.2cm]{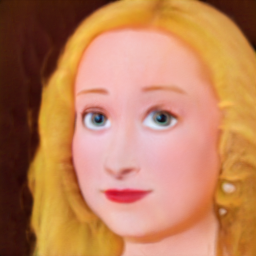}\caption*{\small{(c) face2cartoon}}
\end{subfigure}
\begin{subfigure}{4.2cm}
\includegraphics[width=4.2cm]{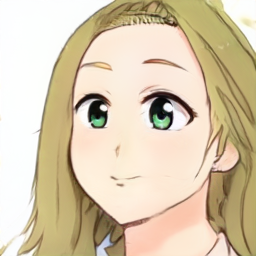}\caption*{\small{(d) cartoon2anime}}
\end{subfigure}
\caption{An example of the proposed multi-domain I2I translation results. We train transformed models on four different face domains and then can perform I2I translation between any two domains, such as portrait2face (b), face2cartoon (c), and cartoon2anime (d). Compared to existing I2I translation methods, which need at least $C_4^2=6$ models to perform arbitrary I2I translation among the above $4$ domains, our method only needs $4$ models since any two of those models can be used for translation.}
\label{fig:exm}
\end{figure*}
Recent work has made huge progress in this field due to the development of the generative adversarial network (GAN) \cite{gan}. After the pioneer work of Pix2Pix \cite{pix2pix}, BicycleGAN \cite{bicyclegan} was developed to support I2I translation with multiple styles. Later, researchers proposed UNIT \cite{unit}, CycleGAN \cite{cyclegan} MUNIT \cite{munit}, and DRIT/DRIT++ \cite{drit, drit++} for unsupervised I2I, where MUNIT and DRIT/DRIT++ can perform multi-modal translation. Nevertheless, there are still challenges in the I2I translation task. First, paired data are limited for supervised training methods while quality of the results of unsupervised methods has been relatively low. In addition, most I2I networks must train at least two generators for source and target domains together. This may lead to some problems: training resources including time and memory, would be high; uneven data in either source or target domains could cause instability in training process. Moreover, controllable generation and multi-domain translation \cite{exemplar, distangle, localcontrolcolor, pirec} attract lots attention in both academia and industry, but most of the proposed methods require additional training data or complex structural design, which makes the original translation much more resource-consuming.\par 

We propose a new solution to the I2I translation task via a pre-trained network. We want to find models that can generate paired images based on the same latent representation. Specifically, for model $G_X$, which can generate image $X\in\mathcal{X}$ based on a given latent code $z$, we try to find the model $G_Y$ that can generate image $Y\in\mathcal{Y}$, where $\mathcal{X}$ and $\mathcal{Y}$ are two image domains, and $G_Y(z)$ is semantically similar to $G_X(z)$. In our initial experiments, we found that with conservative learning methods and proper layer manipulation, fine-tuning model $G_X$ on the $\mathcal{Y}$ domain data can yield model $G_Y$, which satisfies the above assumption. We refer to this process as model transformation $T$ in the following sections, where $G_Y = T_{X\rightarrow Y}(G_X)$. Previous inversion work including IdInvert \cite{indomain} and Image2StyleGAN \cite{image2stylegan} proposed methods to obtain latent representation $z$ of an image $I$ given its pre-trained model $G$. With these method, the I2I problem can be solved by (1) inverting the given image X to obtain the latent representation $z=Inv(X, G_X)$ where $Inv$ is an inversion method; (2) performing model transformation on the pre-trained model $G_X$ to generate the target domain model $G_Y$, and (3) generating the paired image in $\mathcal{Y}$ domain $Y=G_Y(z)$.\par

Different from traditional methods that train from the scratch, we choose pre-trained models like StyleGAN2 \cite{stylegan2} to ensure image quality and fidelity of the output. In addition, our method requires training only in the model translation. This process contains conservative model fine-tuning and layer manipulation, which requires much less training resources in terms of both time and memory. As the model transformation focus only on the given domain, our method is unsupervised. It does not require any paired data across different domains. To achieve unsupervised I2I translation solutions, previous methods including CycleGAN \cite{cyclegan}, MUNIT \cite{munit}, and DRIT/DRIT++ \cite{drit, drit++} all used cycle loss. This means bidirectional translations $X\rightarrow Y$ and $Y\rightarrow X$ should be trained together. Model collapse can happen at one side if the training data are unbalanced. In other words, those methods cannot achieve proper results when data in one domain are limited. Our method that trains models in a single domain at a time can preserve the generation quality as much as possible, and it is not affect by unbalanced data in two domains.\par

Besides, the pre-trained model StyleGAN2 \cite{stylegan2} in nature supports the image generation of multiple styles. By injecting various latent style code at different layers of StyleGAN2 \cite{stylegan2}, the final output can have different styles. This enables our method to support multi-modal I2I translation without any additional training or modification of the network structure. We also found that the proposed method can be extended to multi-domain I2I translation. Assuming that we have a pre-trained model trained on domain $\mathcal{D}_0$, and data from various domains $\mathcal{D}_i$ where $i\neq 0$, with model transformation $T_{D_0\rightarrow D_i}$, we can obtain models for different domains $G_{D_i}=T_{D_0\rightarrow D_i}(G_{D_0}, D_i)$. According to that images generated by $G_{D_0}$ would be semantically similar to images generated by $G_{D_i}$, combined with the transitivity of semantic similarity, such similarities will be preserved between any pairs of two generated models $G_{D_m}$ and $G_{D_n}$ where $m\neq n$. Therefore, as shown in Fig.\ref{fig:exm}, we can perform I2I translation between any two domains $\mathcal{D}_m$ and $\mathcal{D}_n$ without additional training.\par

Our major contributions are:
\begin{itemize}
	\item We define the distance between two models to measure the semantic similarity between two images generated by two models, based on the same input latent vector. The model distance help us analyze the properties of I2I translation between the two domains.
	\item We propose an unsupervised I2I translation method via a pre-trained StyleGAN2 model, which can generate images in the target domain with high quality. Our proposed method can support multi-modal and multi-domain I2I translation. Compared to traditional I2I translation methods, it drastically improved the results and requires much less training resource. 
	\item We also proposed an inversion method, which can perform conversion between an image and a latent vector based on a given StyleGAN2 model. Compared to previous algorithms, our inversion method is based on an embedded GAN space, which provides a boundary constraint for searching the latent code of the input image. 
	\item We perform both qualitative and quantitative evaluations on various challenging dataset, and we compare the proposed method with state-of-the-art works. This shows that the proposed method achieves outstanding output and better results in all the experiment. 
\end{itemize}

%% file: sections/related_work.tex
\subsection{Image-to-Image Translation}
I2I translation aims to transfer an input image from the source domain to target domains. This can be applied to many applications including colorization, super-resolution, and face-editing. With the development of GAN \cite{gan}, Isola et al. proposed the first supervised I2I translation network pix2pix \cite{pix2pix}, which they then extended to pix2pixHD \cite{pix2pixhd} to support translation between images with higher resolution. As the paired training data are limited, an unsupervised method CycleGAN \cite{cyclegan} was then proposed by Zhu et al. They used cycle loss to build up the constraints between the source and target images, which were then applied in other unsupervised I2I translation methods \cite{unsuperdiverse, unsuperemerging, unsupermulti, unsuperXgan}.\par
Moreover, several multi-modal methods including BicycleGAN \cite{bicyclegan}, MUNIT\cite{munit}, and DRIT/DRIT++\cite{drit, drit++} made it possible for an input image to have multiple counterparts with different styles. In those methods, a style latent code could either be randomly sampled or extracted from a given reference image. By injecting the combined content and style code into the generator, they could generate output with specified appearance. However, this type of style control is not accurate, and it can be applied only to images in a global way. Later works \cite{controlexample1, exemplar, tmm_segin} used an example to semantically control the output images, while other works\cite{localcontrolcolor, controlpose, huang2020multi} tried to locally control attributes of the output such as color and pose, based on users' input. Most recently, motivated by the high quality of StyleGAN \cite{stylegan}, some works begin to incorporate it into the I2I framework. Elad et al. \cite{psp} proposed to learn a mapping from images to StyleGAN latent codes via an end-to-end training method, which can achieve high-quality I2I translation between two domains. In addition, some researchers used Toonify \cite{toonify} which is based on the distillation of a blended StyleGAN \cite{stylegan} network into a pix2pixHD \cite{pix2pixhd} I2I translation network. The translation part is similar to the pix2pixHD.\par
Current I2I translation methods require online training on at least two domains, and most of them have domain-distinguished generators and discriminator in an overall network to ensure the output image fits its own domain. However, this has a high demand on training resources in terms of both time and memory, and such demand would increase as the domains increase. Modal collapse also can happen when training data are not balanced in any domains. Therefore, we proposed the method that focus on training images in one domain at a time, and use model transformation to link models trained on different domains to achieve multi-modal and multi-domain I2I translation.

\subsection{Inverse Mapping from Image to Latent Code}
With the rapid development of GAN \cite{gan}, more and more models can generate high quality images from a latent code. These include StyleGAN/StyleGAN2 \cite{stylegan, stylegan2}, BigGAN \cite{biggan} and PCGAN \cite{pcgan}. Researchers further studied the GAN space and tried to find a way to map between the input latent code and the output image. Perarnau et al. proposed IcGAN \cite{icgan}, which trained an encoder in the cGAN framework to compress a real image $X$ into a latent representation $z$. Similarly, AEGAN \cite{aegan} was then proposed by Luo et al. It used an autoencoder to train an inverse model by minimizing the differences between noise-generated samples and related reconstructed images. Ma et al. \cite{invertibility} theoretically proved that a convolutional GAN is invertible that same inversion could be achieved even if only a subset of samples is observed. \par

Then researchers found that it was also possible to use optimization methods \cite{invertinggan, precise_recovery,image2stylegan,image2stylegan++} to search the latent representation iteratively, according to their experiments, the optimization-based methods could provide more accurate noise as well as better reconstructed images. Gu et al.\cite{mganpiror} believed that one image could be determined by multiple priors. They proposed the Multi-code GAN prior for image inversion. They proposed that multiple latent representation could reconstruct the image with more details and higher precision. To speed up the inversion process, Zhu et al. \cite{indomain} proposed the in-domain GAN inversion method. Combining the optimisation with a trained encoder, the inversion encoder can output a rough latent representation which could then be taken as an initial vector in the optimization. This could greatly diminish the optimisation time without losing the precision of the inversion.
\par

Most inversion methods use the difference between the input image and the reconstructed images as major loss functions. Such difference includes general L1/L2 loss and perceptual loss. In addition to those loss functions, there are no extra constraints applied to the latent representation or latent space. Our proposed method adds an explicit boundary to 
the latent space so that the output latent code can always be a point in the pre-trained GAN space. \par
\subsection{Transfer Learning}
Transfer learning aims to solve problems using data that are not directly related to the task considered. This can be classified as two major types: (1) similar data for different tasks or (2) different dataset for the same task. The second type tries to find a solution for the task with limited data. Our method can also be classified into the second type as it aims to perform I2I translation (same task) among images in multiple domains (different dataset). Transfer learning contains many typical learning methods such as conservative learning, multi-task training, domain adversarial learning and teacher-student systems. Domain adversarial learning and multi-task training focus on solving the first type of problem, while conservative learning is proposed to solve the limited-data learning problems. Conservative learning work including \cite{TL_conserve1, TL_conserve2, TL_converve3, TL_ft1,TL_ft2, TL_ft3, TL_ft4, TL_ft5} conducts fine-tuning with partial layers, while other works \cite{TL_r1, TL_r2, TL_r3, TL_r4} add a regularizator or loss functions to constraint the distance between the current output and the pre-trained output. Since the teacher-student structure is an essential subject in the transfer learning system, we do not discuss it here because it is out of the scope of this work.\par

%% file: sections/method.tex
\begin{figure*}[!ht]
   \centering
\includegraphics[width=\textwidth]{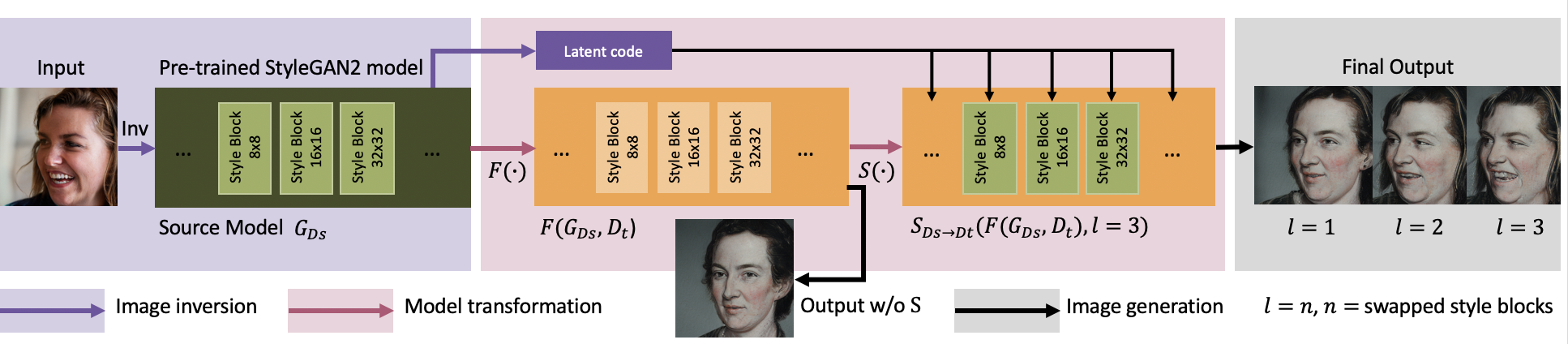}
\caption{The main structure of the proposed work. The left-most block aims to obtain a latent code from a given image via the inversion method. And the middle part is the model transformation, where $F$ stands for fine-tuning process and $S$ represents high-level layer-swap. Although results generated by the model after $F$ process belongs to the target domain, later $S$ transformation can achieve higher semantic similarities to the input images and such similarities increase with the increase of the number of swapped layers.}
\label{fig:framework}
\end{figure*}

To describe the semantic similarities between images generated by different models, we define the model distance in a formal way as a metric to measure such similarities. Then we propose a series of transformations $T$ to reduce the distance between models trained on different domains. We also propose an improved inversion method to obtain the latent code from images; This enables I2I translation between semantically similar models. As we use StyleGAN2 \cite{stylegan2} as the base model, our method not only ensures high-quality results; it also supports multi-modal translation in its essence. The overall flow of the proposed method is shown in Fig. \ref{fig:framework} \par
\subsection{Model distance and transformation}
The I2I translation task is based on deep convolution neural network (DCNN). With an input image, the network can generate a related output in the target domain. Deep neural models are determined by model structures and parameters, since there are no unique criteria for structural design, and the amount of parameters in one model could be huge. Therefore, it is difficult to evaluate the relationship between two models directly. Although this seems like a black box problem, we can define the distance from another point of view. That is, we can use the distance between two outputs from the models given the same input as a representation of the model distance. Two models, $G_{D_i}$ and $G_{D_j}$, can generate images from latent code $z$ in two domains $D_i \in \mathcal{D}_i$ and $D_j \in \mathcal{D}_j$. The model distance between $G_{D_i}$ and $G_{D_j}$ can be defined as:
\begin{equation}
    d(G_{D_i}, G_{D_j}) = \int dist(G_{D_i}(z), G_{D_j}(z))dz, z\in \mathcal{N}(0, \mathcal{I})
    \label{eqn:dist}
\end{equation}
where $dist$ is the semantic similarity between two images. Specifically, we use LPIPS \cite{lpips} distance in the proposed method.\par 
Based on the previous discussion, the I2I translation can achieve reasonable results only if there is a small distance between the models in the source and target domains. Otherwise, the latent code obtained from the source image $z=Inv(D_i, G_{D_i})$ cannot generate a related output in the target domain. That is, $D_i \nsim D_j$, where $\sim$ indicates high semantic similarity between two images. In our experiments, we found fine-tuning can reduce the model distance. We use the StyleGAN2 \cite{stylegan2} model trained on source domain$\mathcal{D}_s$ as the base model, and we fine-tune it with data in target domain $\mathcal{D}_t$ with the GAN loss:
\begin{equation}
    \begin{split}
    \mathcal{L}_{adv}(G, D) &= \mathbb{E}_{D_t\sim \mathcal{D}_t}[logD(D_t)]\\ 
    &+ \mathbb{E}_{z\sim\mathcal{N}}[log(1-D(G_{D_t}(z))] 
    \end{split} 
\end{equation}
As we know, the StyleGAN2 network first maps the latent code $z$ to an embedded space $\mathcal{W}$ via an 8-layer MLP $w=f(z), w\in\mathcal{W}$, where $f:\mathcal{N}\rightarrow \mathcal{W}$ is the feature mapping function. Features in $\mathcal{W}$ space are then be sent to the generator to output the images. Inspired by FreezeD \cite{freezed}, we freeze the fully-connected (FC) layers in the fine-tuning process, in which the fine-tuned model can achieve smaller model distance from the base model. This is because, with a frozen FC layers, the fine-tuned model would have the same embedded space as the base model. This ensures that the input latent code is always initialized at the same point in the embedded space as that of the base model. \par

After that, given a latent code $z$, the fine-tuned model $G_{D_t} = F(G_{D_s}, \mathcal{D}_t)$, where $F$ indicate the fine-tuning process, can generate images $D_t=G_{D_t}(z), D_t\in\mathcal{D}_t$ and $D_t\sim G_{D_s}(z)$. However, during fine-tuning, the semantic similarity decreases due to domain difference. The new model $G_{D_t}$ must learn specific features in the new domain to reduce the GAN loss. This is similar to catastrophic forgetting \cite{forget} in transfer learning where the new model $G_{D_t}$ would fail to generate images in the source domain after fine-tuning, and it forgets partial features previously learned from the source domain. Inspired by Justin et al. \cite{justin}, they applied layer-swapping on image analogy task. And we found it is possible to swap higher convolutional layers of $G_{D_s}$ to $G_{D_t}$. Then the target model can generate images and preserve more features from the source domain. This further diminishes the model distance between $G_{D_s}$ and $G_{D_t}$. We define this process as $G_{D_t} = S_{D_s\rightarrow D_t}(G_{D_t}', l), l\in N$, where $G_{D_t}'$ is the model after fine-tuning and $l$ indicates the number of swapped layers. According to experiments, we found layer-swap cannot be used between two models with large model distance (e.g. models without fine-tuning). This is because features in such models have large differences, they might even be two feature spaces. Therefore features outputted by swapped layers cannot be used by the following layers. However, we can control the final output of $G_{D_t}$ by modifying the number of swapped layers $l$. With a larger $l$, the generated images are more semantically similar to the images in the source domain. With a smaller $l$, the generated images maintain more features in the target domain. Hence, compared to freezing the generator directly in the fine-tuning stage, layer-swap provides more flexibility in controlling the final output. Finally, We define the transformation for reducing the model distance between $G_{D_s}$ and $G_{D_t}$ as 
\begin{equation}
    G_{D_t} = T_{D_s\rightarrow D_t}(G_{D_s}) = S_{D_s\rightarrow D_t}(F(G_{D_s}, \mathcal{D}_t), l)
\end{equation}
With model transformation, models with short distances can be created in various domains.
\subsection{Image-to-code inversion}
So far, we have two models in different domains, where $G_{D_s}(z) \sim G_{D_t}(z)$. Specifically, StyleGAN has an MLP mapping the latent code $z$ to an embedded space, that is $G_{D_s}(w)\sim G_{D_t}(w)$ where $w=f(z)$. However, in the I2I translation task, the input is an image in the source domain $I_s$, and the required output $I_t$ should be corresponding images in the target domain, namely $I_s \sim I_t$. To achieve the I2I translation, we need to bridge the gap between the input image and the embedded latent vector $w$. Therefore, we propose an inversion method to find the latent code related to the input image $w = Inv(D_s, G_{D_s})$. The latent code will then be sent to the transformed model to generate output in the target domain $D_t = G_{D_t}(w)$. This process is defined as:
\begin{equation}
    D_t = T_{D_s\rightarrow D_t}(G_{D_s})(Inv(D_s, G_{D_s}))
\end{equation}
where $Inv$ is the inversion algorithm taking an image and a related model trained in the same domain as the input.\par
An algorithm to project the image to latent space has been proposed in StyleGAN2, which obtains the latent code directly by optimizing loss functions:
\begin{equation}
    w^* = \argmin_{w\in \mathbb{R}^d}\quad ||I-G(w)||_1 + ||\Phi(I)-\Phi(G(w))||_2
    \label{eqn:opt}
\end{equation}
where $I$ is the input image, $G$ represents the pre-trained network on the related domain and $\Phi$ indicates the pre-trained VGG \cite{vgg} network, which is used to extract image features.\par
This method tries to find a specific $w$ in the embedded space $\mathcal{W}$, but it cannot guarantee the found $w$ is within the embedded space for the following reasons: (1) there is no explicit description about the $\mathcal{W}$ space, as it is formed by fully connected layers. (2) there is no constraint directly applied to $w$, as the loss functions are based on images. To tackle this problem, we propose an improved method that optimises the $w$ within an explicit defined embedded space. Shen et al. \cite{closedform} proposed SeFa to find a semantic space for a pre-trained GAN model, where major directions in the space can control specific semantic attributes in generated images. In SeFa \cite{closedform}, the difference between two latent codes can be described as:
\begin{equation}
    \Delta w=(Az+b) - (A(z+\alpha n)+b) = \alpha An
\end{equation}
where $A\in \mathbb{R}^{m\times d}$, $b\in \mathbb{R}^d$ are weight and bias in the fully-connect layers and $\alpha$ is a hyperparameter. SeFa states that $A$ is a semantic selector of the model, and finding a specific semantic attribute is to optimize $n^*=\argmax ||An||^2_2$, and $n^*$ is the direction for that attribute. This can then be extended to a problem finding the top k major semantic directions, so the optimization would be rewritten as:
\begin{equation}
    N^* = \argmax_{N\in \mathbb{R}^{d\times k}} \sum^k_{i=1}||An_i||^2_2-\sum^k_{i=1}\lambda_i(n^T_in_i -1)
\end{equation}
where $N=[n_1, n_2, ..., n_k]$ and $\sum^k_{i=1}\lambda_i(n^T_in_i -1)$ is the Lagrange Multiplier. After taking the partial derivative on $n_i$, we have:
\begin{equation}
    A^TAn_i-\lambda_in_i=0
\end{equation}
According to SeFa \cite{closedform}, the eigen vectors of $A^TA$ are semantic directions. Therefore we can use this metric as a formed embedded space $V=A^TA$. We rewrite the optimization in Eq.\ref{eqn:opt} as follows:
\begin{equation}
    v^* = \argmin_{v\in \mathbb{R}^d}\quad ||I-G(v^TV)||_1 + ||\Phi(I)-\Phi(G(v^TV))||_2
    \label{eqn:opt2}
\end{equation}
where $v$ is a projection point in space $V$, and the latent vector $w=v^TV$. Specifically, we first compute $V_{D_s}$ based on a pre-trained model in the source domain $G_{D_s}$, and we obtain the vector $v_{D_s}$ by optimizing Eq.\ref{eqn:opt2}. Then we have the inverted latent code for a given input image $D_s$ that $w=v_{D_s}^TV_{D_s}=Inv(D_s, G_{D_s})$. Compared to direct projection, our proposed method is more like selecting one point in a pre-defined embedded space, based on a given network. The constraints of latent space can improve the previous optimization. 
\subsection{Multi-modal and Multi-domain I2I translation}
Similar to MUNIT \cite{munit} and DRIT/DRIT++ \cite{drit,drit++}, the proposed method can support multi-modal I2I translation as well. First, we revisit the StyleGAN2 \cite{stylegan2} structure that latent code will be mapped to layer-wise style code and then sent to the major generator via AdaIn \cite{adain} layers. StyleGAN \cite{stylegan} states that style code injected at the higher layers can change a major structure of the output, such as identity, while the style code applied at lower layers can modify only minor features such as color, light conditions and other micro structures.\par
In our method, we separate the $w$ into two parts, namely content code $w_c$ and appearance code $w_a$, and $w={w_c, w_a}$. To ensure high semantic similarities between the source and the target image, $w_c$ is obtained via the inversion method from the source image $w_c=Inv(D_s, G_{D_s})$, while the appearance code $w_a$ can be sampled from normal distribution $z_a\in \mathcal{N}(0, \mathcal{I}), w_a=f(z_a)$. To perform multi-modal I2I translation, $w_c$ is inserted into most generator layers, while $w_a$ is used at only the last few layers, the details of implementation are in Section IV. Therefore, multiple output in the target domain can be generated based on the input:
\begin{equation}
    D_t^{(i)} = G_{D_t}(Inv(D_s, G_{D_s}), w_a^{(i)})\quad i=\{1, 2, ..., n\}
\end{equation}
Moreover, the appearance code can also be obtained from a reference image from the target domain $w_a=Inv(D_r, G_{D_t}), D_r\in \mathcal{D}_t$. Similar to layer-swap mentioned above, we can choose different numbers of layers to inject the appearance code to control the similarities between the generated output and the reference.\par
In addition, the proposed method can support multi-domain translation. In traditional methods, multi-domain translation is limited because it requires that multiple generators are trained together. This places a high demand on both resources and data quality. All the training in our method focuses one domain at a time, and multi-domain translation can be performed off-line with pre-trained models. Assuming that we have an initial source domain data $\mathcal{D}_s$ and a bunch of target domain data $\{\mathcal{D}_{t1}, \mathcal{D}_{t2}, ..., \mathcal{D}_{tn}\}$, according to the proposed transformation method $T_{D_s\rightarrow D_{ti}}$, we can obtain models $G_{D_{ti}}$ in such domains. Let $\delta = d(G_{D_s}, G_{D_{ti}})$, where $\delta$ is a small value and the semantic similarity has a certain degree of transitivity. For instance, if A is semantically similar to both B and C, it is impossible that B is totally irrelevant to C. Therefore, we have $d(G_{D_{ti}}, G_{D_{tj}}) \simeq \delta$. In that way, models in different domains $D_{ti}$, and $D_{tj}$ satisfy that $G_{D_{ti}}(z) \sim G_{D_{tj}}(z)$ and we can perform I2I translation between any pair of domains in the given target domains.\par
In the proposed method, the training process is required only in the model transformation, and it focuses on a single domain. This greatly reduces computational cost and the amount of overall memory used. Our experiments show that even with a small dataset, and conservative tuning could still achieve satisfactory results and the fine-tuned model could generate images with high quality and fidelity.

%% file: sections/experiments.tex
To evaluate the proposed I2I translation method, we conducted experiments on various datasets for the I2I translation task. We also compared the proposed methods with several state-of-the-art work, including CycleGAN \cite{cyclegan}, MUNIT \cite{munit} and DRIT++ \cite{drit}. Both the qualitative and quantitative results show that the proposed method can generate images in the target domains with much higher quality. Moreover, the proposed method also preserves the similarity between the input and output images to a very large extent. Besides, experiments show that the proposed method can perform the I2I translation task with multi-modal results in multiple domains. The evaluation of the proposed inversion algorithm and the ablation studies are discussed in detail in the following section. 
\subsection{Baseline and Dataset}
Several state-of-the-art I2I translation work were chosen as baselines for evaluation. CycleGAN is the first unsupervised I2I translation network that proposed cycle consistency loss to ensure identity invariance between the input and output images. Although it does not require any paired data, it could support only single modal translation. MUNIT \cite{munit} and DRIT++ \cite{drit++} are proposed later to support multi-modal translation with better quality. Both methods assumed that images could be encoded into content and appearance latent space and that recombining the content code with various style codes in the target domain could achieve I2I translation. \par
We evaluated the proposed methods in different I2I translation scenarios including face2portrait, face2cartoon, face2anime, cat2dog and cat2wild. We used FFHQ \cite{ffhq} for all the face cases and AFHQ for cats, dogs and wild animals. As for the portrait, anime and cartoon, we used dataset \cite{portrait}, Danbooru2018 \cite{danbooru2018} and \cite{toonify}, respectively. Among those tasks, the face2anime I2I translation was more relatively challenging as the domain distance between the human face and anime data is quite large.\par
\subsection{Implementation Details}
For all the A2B cases, we froze the FC part of the A model, and we fine-tuned model A with data in the B domain. The fine-tuning contained around $12\,000$ (low resolution cases) - $20\,000$ (high resolution cases) iterations. We kept the training strategy and loss functions the same as those in the original StyleGAN2. Specifically, we used 1024x1024 images to fine-tune the pre-trained face model on the FFHQ dataset in face2portrait and face2cartoon. The face2anime model and the other animal related cases were fine-tuned with 256x256 images. In the layer-swap process, we found there are several convolution layers (8x8 - 32x32) in the styleGAN2 network to control the major content features of images in a domain. All of them have 512 channels. As we mentioned before, $G_{D_t}=S_{D_s\rightarrow D_t}(G'_{D_t},l)$. We used $l=1,2,3$ to indicate the layers with size 8x8, 16x16 and 32x32, respectively. Moreover, Adam optimizer was adopted in the proposed inversion method and 1000 iterations were required for all the inversion tasks. The fine-tuning process was conducted on a 2080Ti GPU with 11G memory. The fine-tuning took around 2 days. In addition, it would take around $0.3\,ms$ for layer-swap and model-apply to generate an images in the target domain, and another $0.8-1\,s$ for the inversion process.
\subsection{Qualitative Results}
\begin{figure*}[!ht]
   \centering
%---------row 1
\begin{subfigure}{2.8cm}
\includegraphics[width=2.8cm]{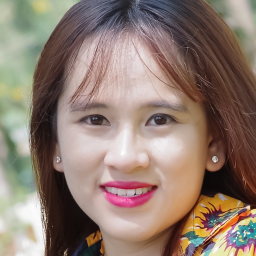}
\end{subfigure}
\begin{subfigure}{2.8cm}
\includegraphics[width=2.8cm]{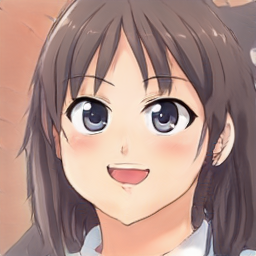}
\end{subfigure}
\begin{subfigure}{2.8cm}
\includegraphics[width=2.8cm]{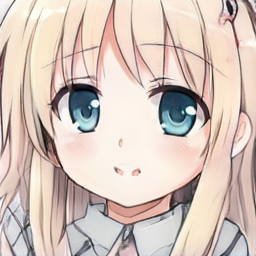}
\end{subfigure}
\begin{subfigure}{2.8cm}
\includegraphics[width=2.8cm]{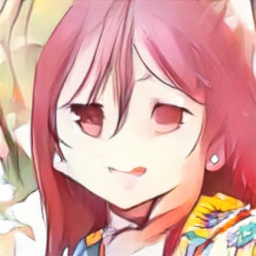}
\end{subfigure}
\begin{subfigure}{2.8cm}
\includegraphics[width=2.8cm]{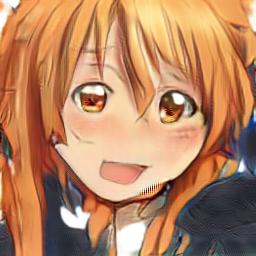}
\end{subfigure}
\begin{subfigure}{2.8cm}
\includegraphics[width=2.8cm]{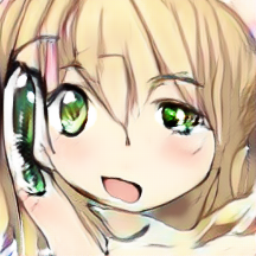}
\end{subfigure}

\begin{subfigure}{2.8cm}
\includegraphics[width=2.8cm]{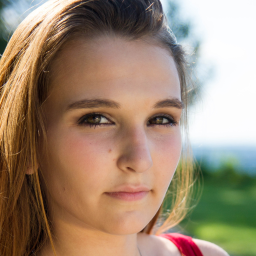}
\end{subfigure}
\begin{subfigure}{2.8cm}
\includegraphics[width=2.8cm]{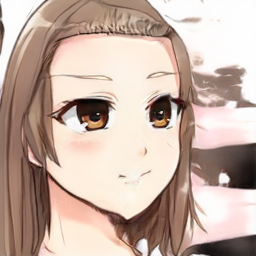}
\end{subfigure}
\begin{subfigure}{2.8cm}
\includegraphics[width=2.8cm]{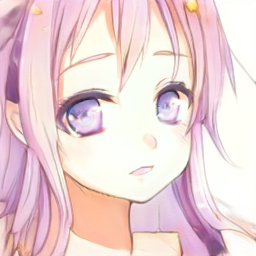}
\end{subfigure}
\begin{subfigure}{2.8cm}
\includegraphics[width=2.8cm]{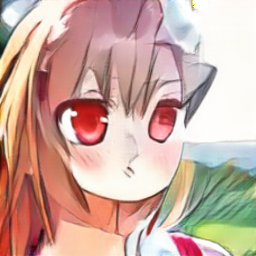}
\end{subfigure}
\begin{subfigure}{2.8cm}
\includegraphics[width=2.8cm]{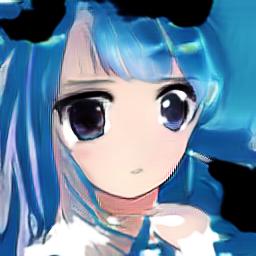}
\end{subfigure}
\begin{subfigure}{2.8cm}
\includegraphics[width=2.8cm]{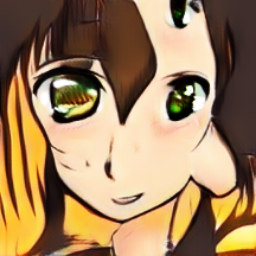}
\end{subfigure}

\begin{subfigure}{2.8cm}
\includegraphics[width=2.8cm]{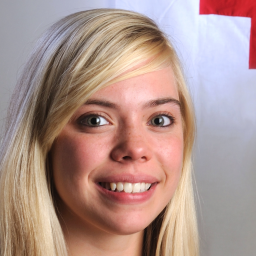}
\end{subfigure}
\begin{subfigure}{2.8cm}
\includegraphics[width=2.8cm]{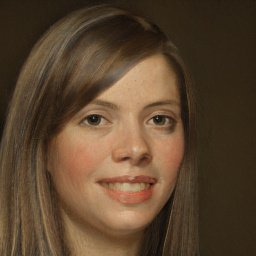}
\end{subfigure}
\begin{subfigure}{2.8cm}
\includegraphics[width=2.8cm]{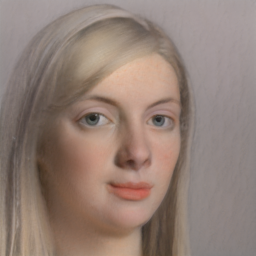}
\end{subfigure}
\begin{subfigure}{2.8cm}
\includegraphics[width=2.8cm]{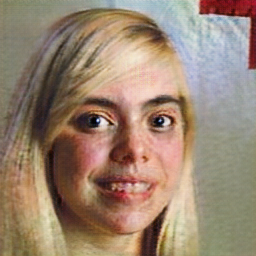}
\end{subfigure}
\begin{subfigure}{2.8cm}
\includegraphics[width=2.8cm]{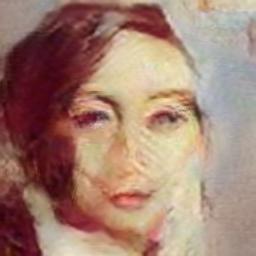}
\end{subfigure}
\begin{subfigure}{2.8cm}
\includegraphics[width=2.8cm]{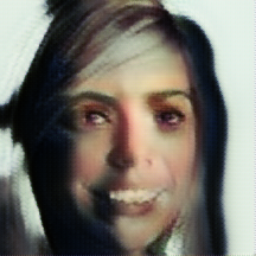}
\end{subfigure}

\begin{subfigure}{2.8cm}
\includegraphics[width=2.8cm]{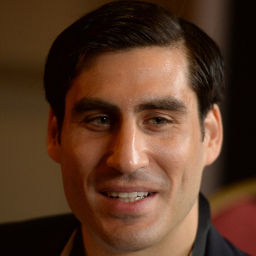}\caption*{\small{Intput}}
\end{subfigure}
\begin{subfigure}{2.8cm}
\includegraphics[width=2.8cm]{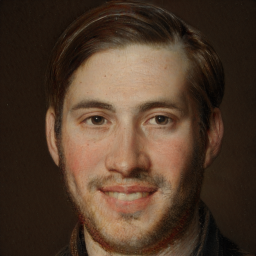}\caption*{\small{Ours}}
\end{subfigure}
\begin{subfigure}{2.8cm}
\includegraphics[width=2.8cm]{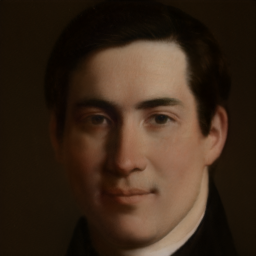}\caption*{\small{Ours w/o LS}}
\end{subfigure}
\begin{subfigure}{2.8cm}
\includegraphics[width=2.8cm]{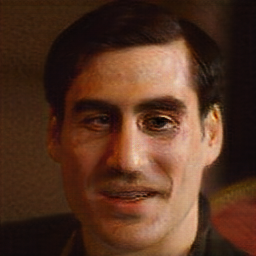}\caption*{\small{CycleGAN \cite{cyclegan}}}
\end{subfigure}
\begin{subfigure}{2.8cm}
\includegraphics[width=2.8cm]{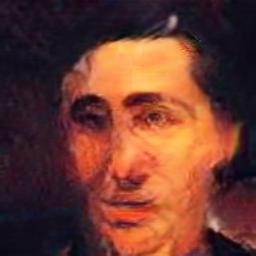}\caption*{\small{MUNIT \cite{munit}}}
\end{subfigure}
\begin{subfigure}{2.8cm}
\includegraphics[width=2.8cm]{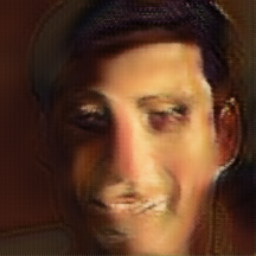}\caption*{\small{DRIT++ \cite{drit++}}}
\end{subfigure}
\caption{Comparison with state-of-the-art work. The proposed method generates images with much higher quality, and it also preserves the semantic characteristics of the input image. It can be observed that the output with LS (the second row) is more semantically similar to the input than the output without LS (the third row)}
\label{fig:comp_all}
\end{figure*}

\begin{figure*}[!ht]
   \centering
%---------row 1
\begin{subfigure}{2.8cm}
\includegraphics[width=2.8cm]{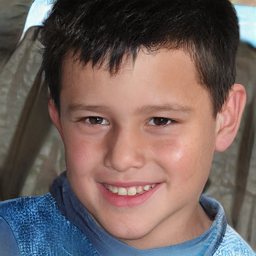}
\end{subfigure}
\begin{subfigure}{2.8cm}
\includegraphics[width=2.8cm]{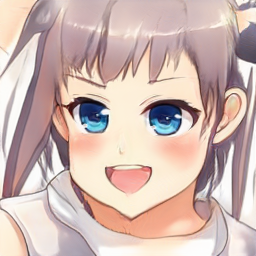}
\end{subfigure}
\begin{subfigure}{2.8cm}
\includegraphics[width=2.8cm]{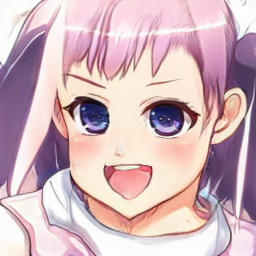}
\end{subfigure}
\begin{subfigure}{2.8cm}
\includegraphics[width=2.8cm]{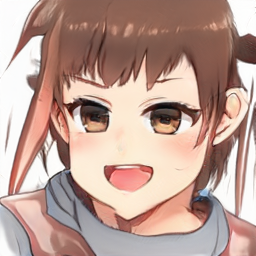}
\end{subfigure}
\begin{subfigure}{2.8cm}
\includegraphics[width=2.8cm]{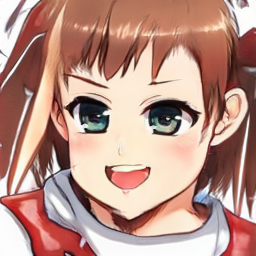}
\end{subfigure}
\begin{subfigure}{2.8cm}
\includegraphics[width=2.8cm]{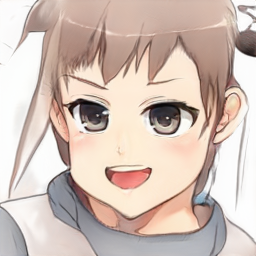}
\end{subfigure}

%---------row 2
\begin{subfigure}{2.8cm}
\includegraphics[width=2.8cm]{figs/fig_multi/anime8009268845030607599_a.png}
\end{subfigure}
\begin{subfigure}{2.8cm}
\includegraphics[width=2.8cm]{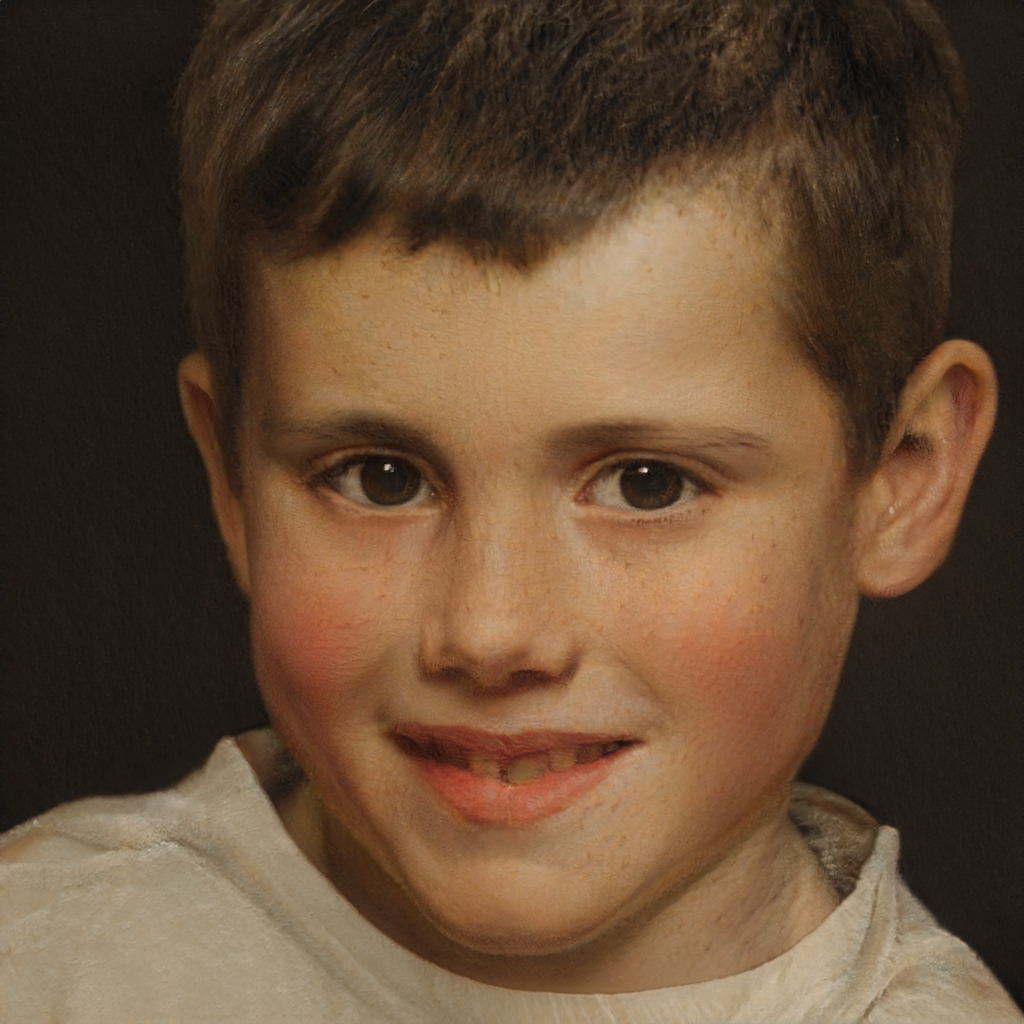}
\end{subfigure}
\begin{subfigure}{2.8cm}
\includegraphics[width=2.8cm]{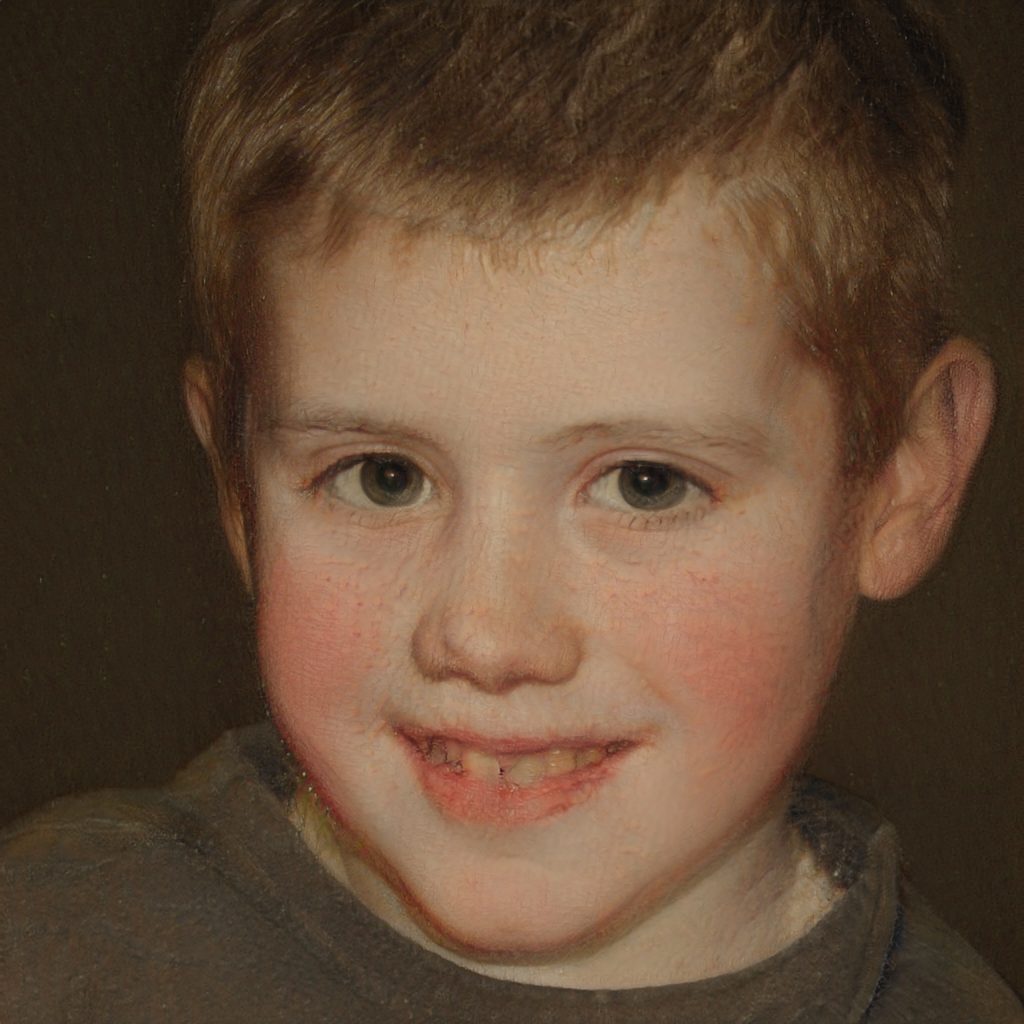}
\end{subfigure}
\begin{subfigure}{2.8cm}
\includegraphics[width=2.8cm]{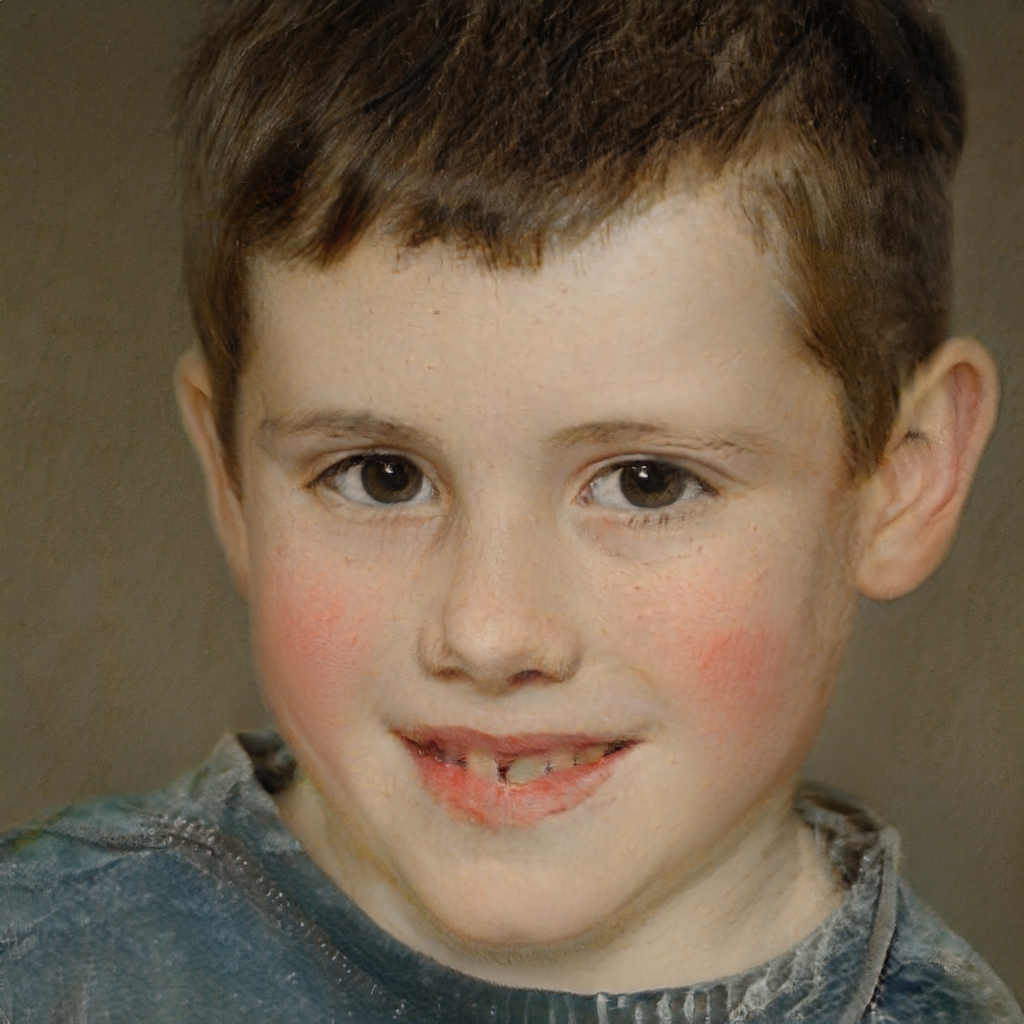}
\end{subfigure}
\begin{subfigure}{2.8cm}
\includegraphics[width=2.8cm]{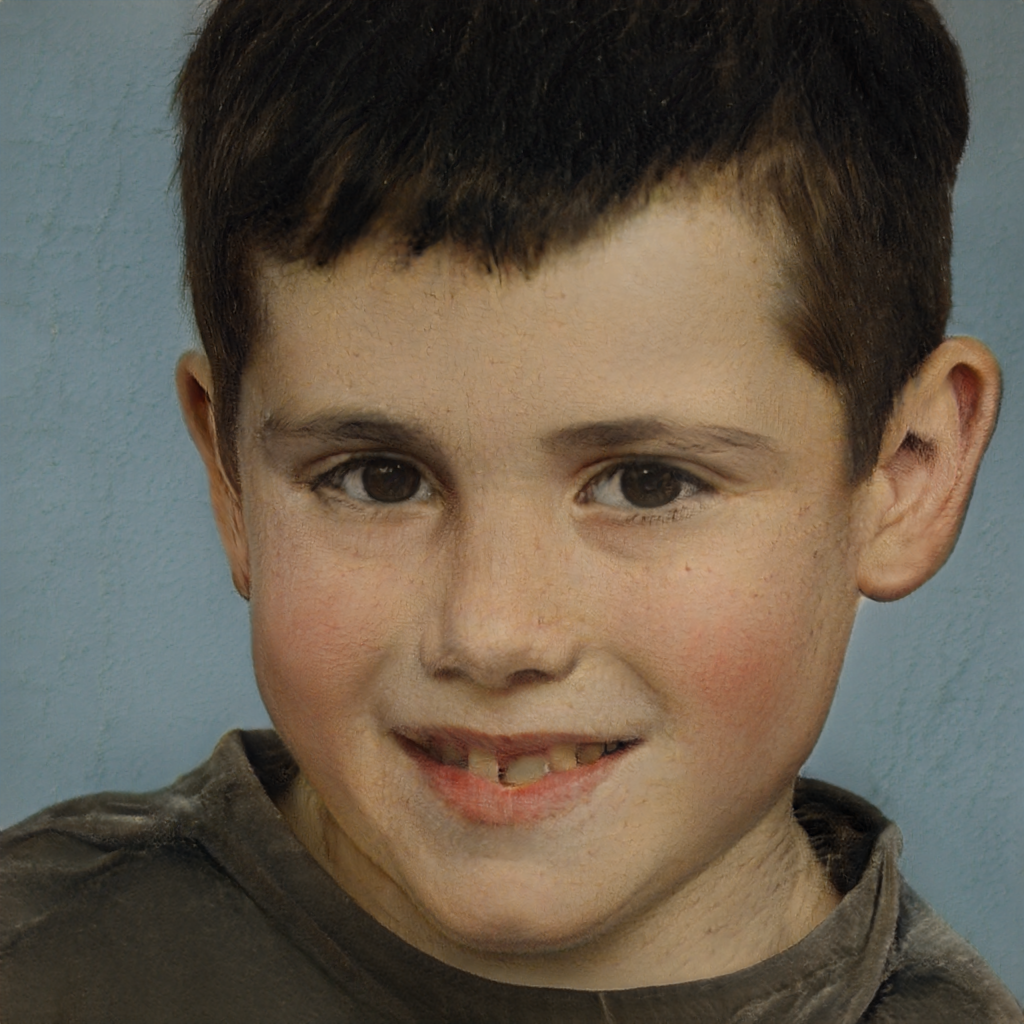}
\end{subfigure}
\begin{subfigure}{2.8cm}
\includegraphics[width=2.8cm]{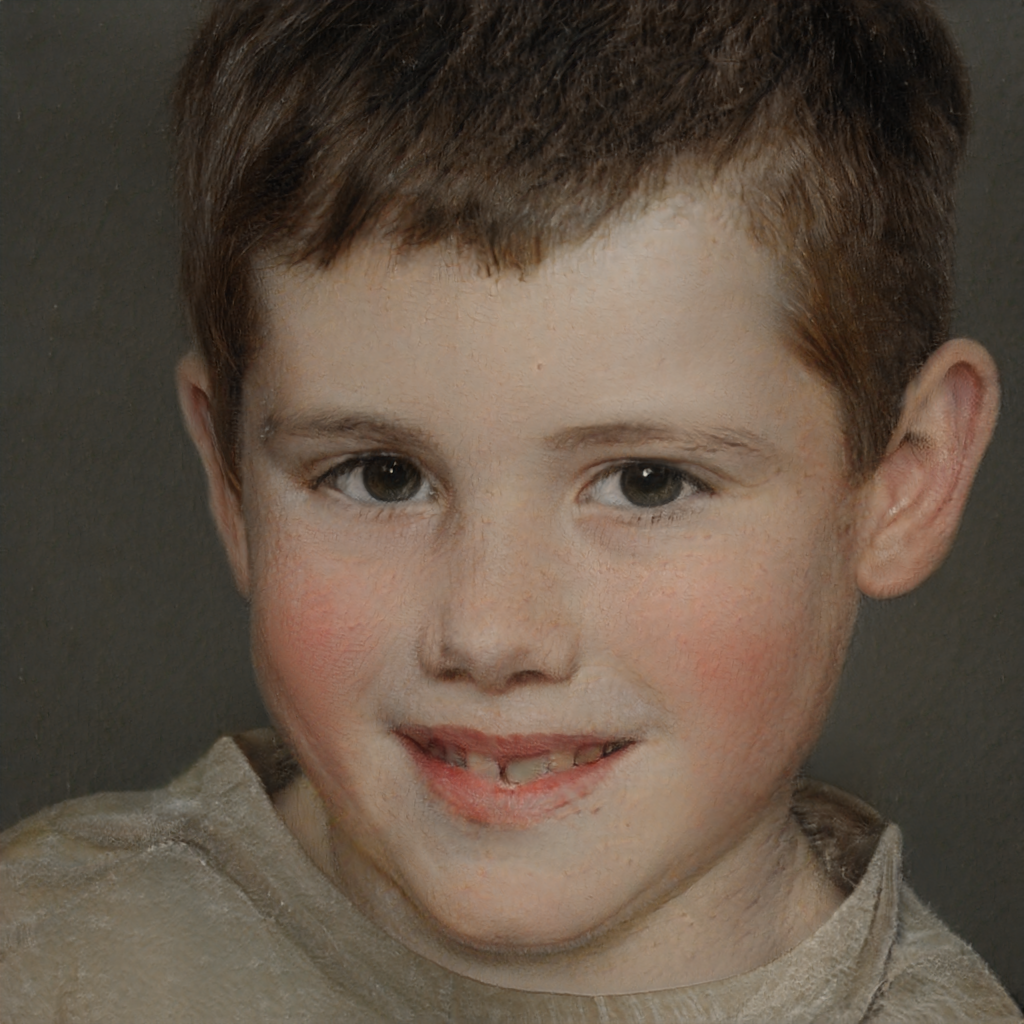}
\end{subfigure}

%---------row 3
\begin{subfigure}{2.8cm}
\includegraphics[width=2.8cm]{figs/fig_multi/anime8009268845030607599_a.png}\caption*{\small{Source domain}}
\end{subfigure}
\begin{subfigure}{2.8cm}
\includegraphics[width=2.8cm]{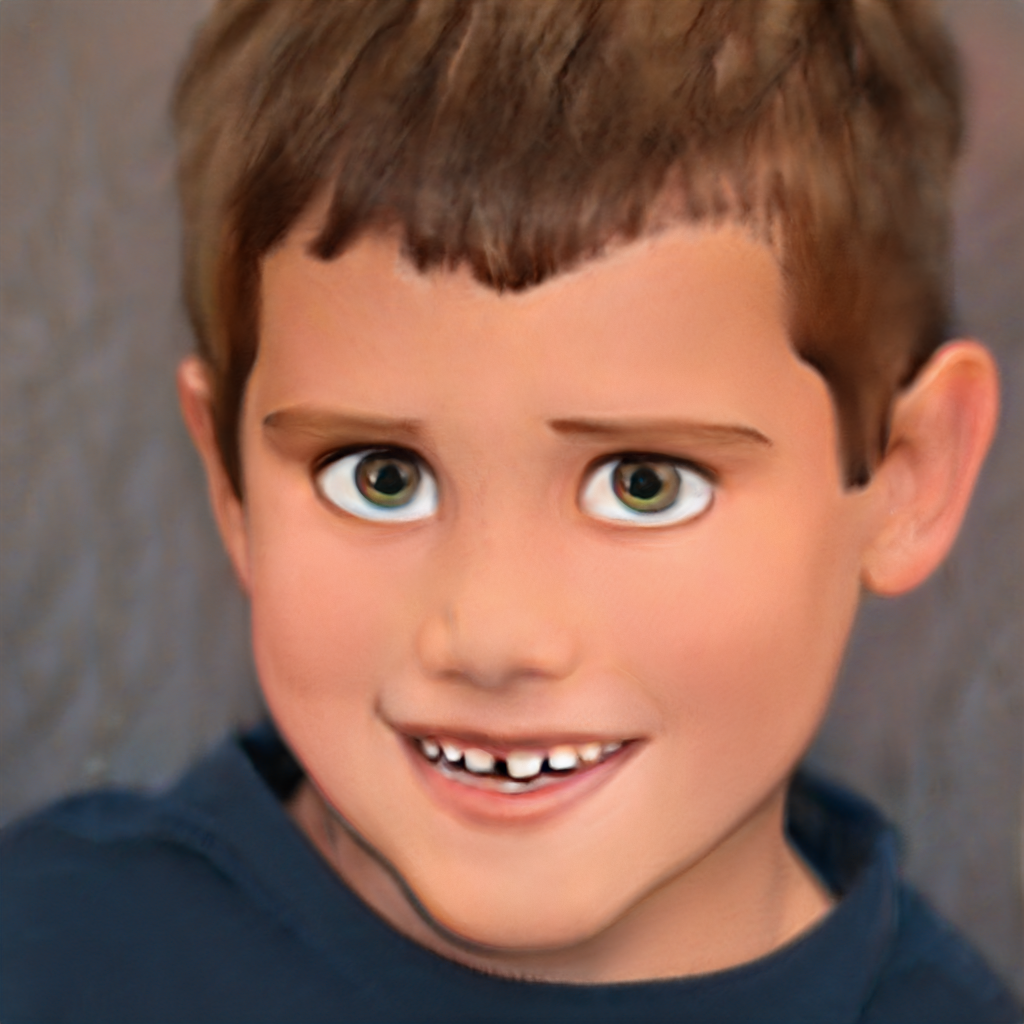}\caption*{\small{Style 1}}
\end{subfigure}
\begin{subfigure}{2.8cm}
\includegraphics[width=2.8cm]{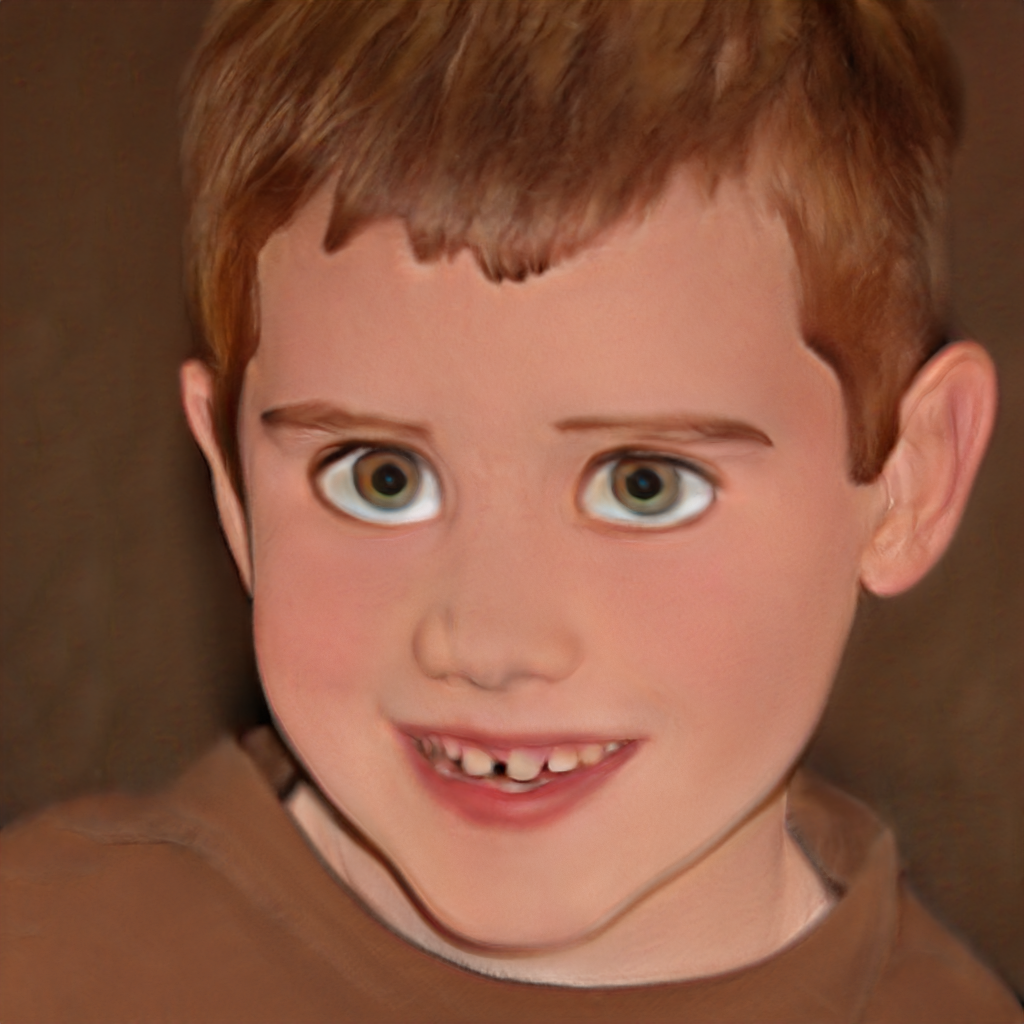}\caption*{\small{Style 2}}
\end{subfigure}
\begin{subfigure}{2.8cm}
\includegraphics[width=2.8cm]{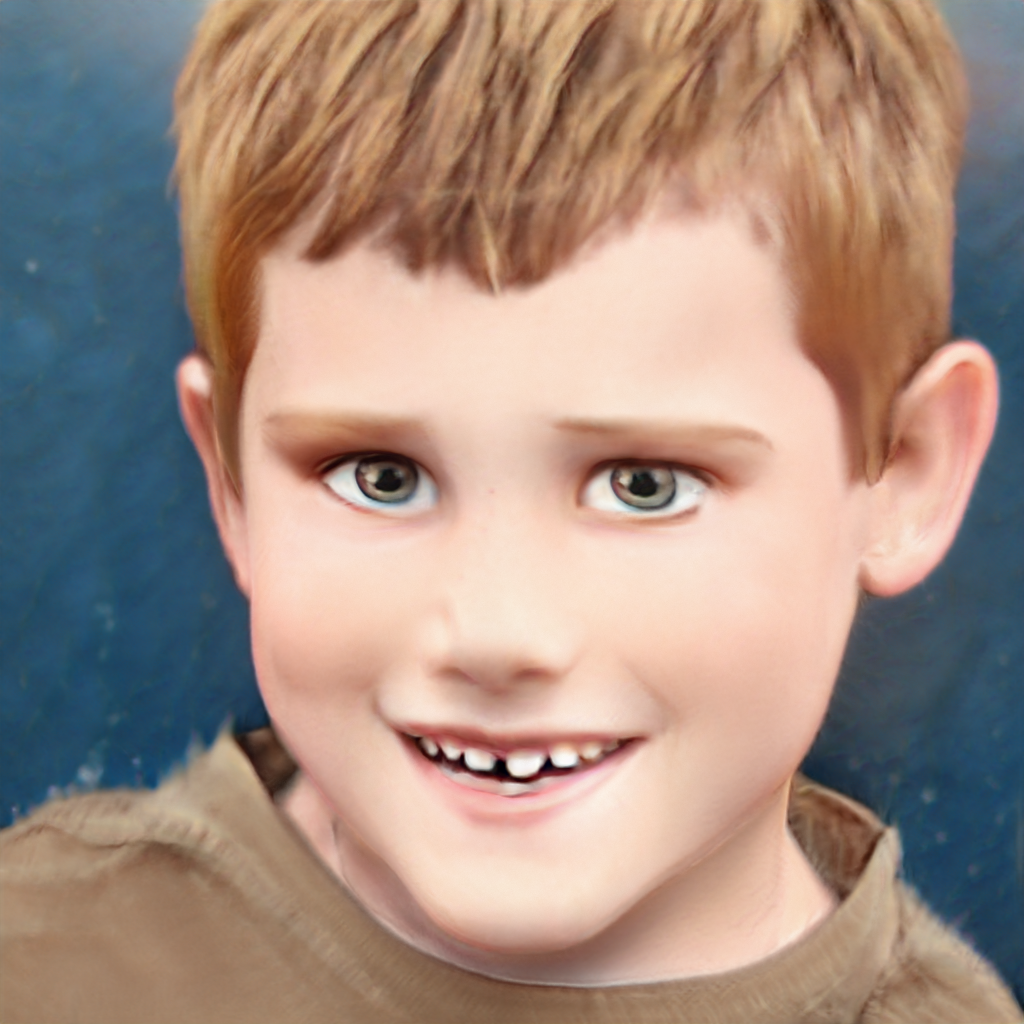}\caption*{\small{Style 3}}
\end{subfigure}
\begin{subfigure}{2.8cm}
\includegraphics[width=2.8cm]{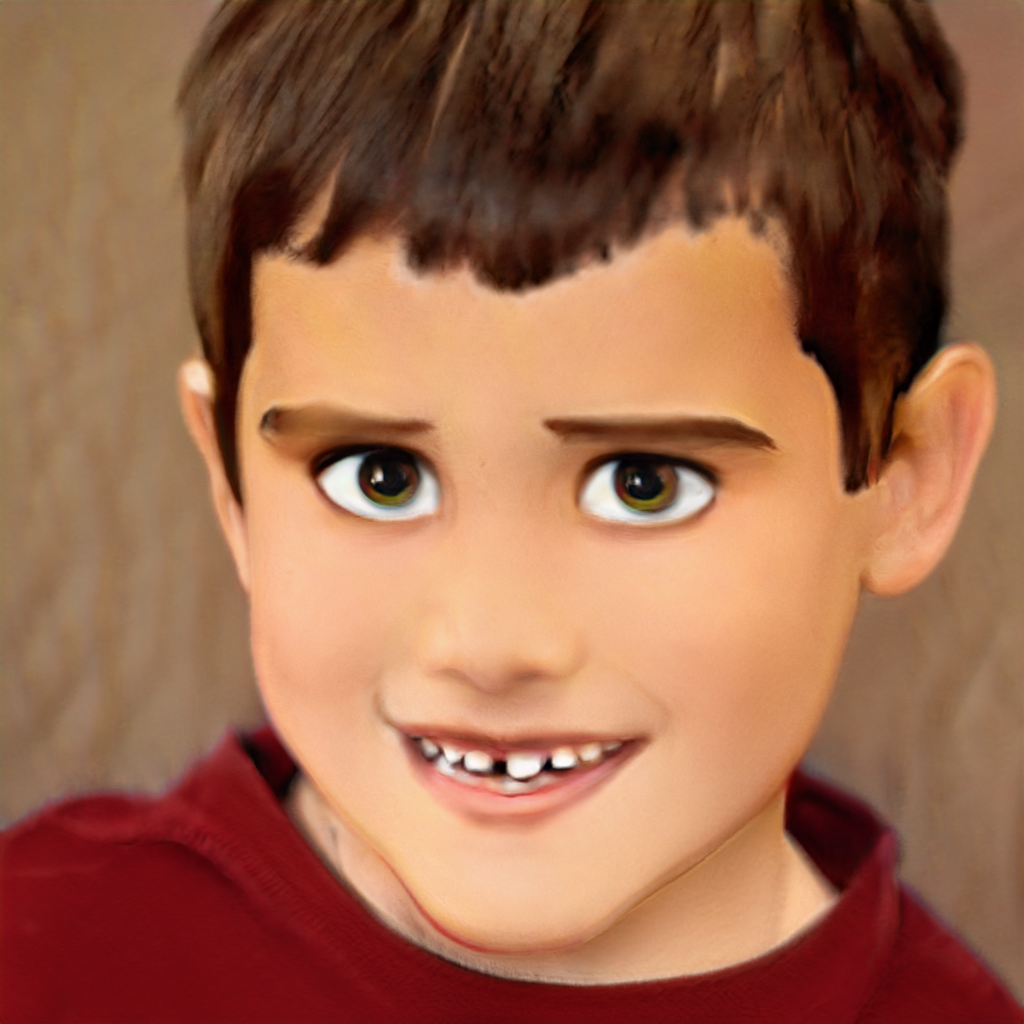}\caption*{\small{Style 4}}
\end{subfigure}
\begin{subfigure}{2.8cm}
\includegraphics[width=2.8cm]{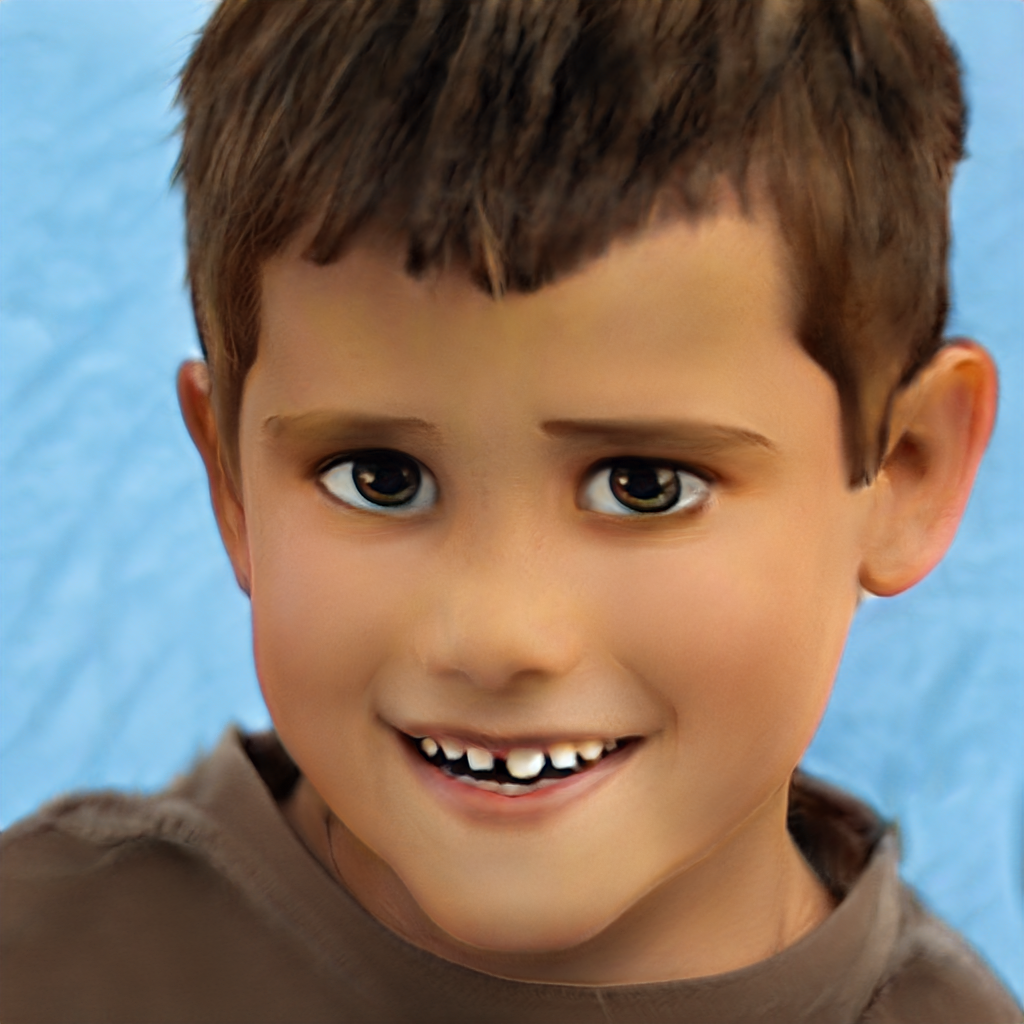}\caption*{\small{Style 5}}
\end{subfigure}
\caption{Multi-modal and Multi-domain translation. Based on the same source, the transformed models can generate corresponding images with different styles in the target domains. I2I translation can be performed between any two of those domains. }
\label{fig:multi}
\end{figure*}

\begin{figure*}[!ht]
%---------row 1
%\hspace{0.01cm}
\rotatebox[origin=c]{90}{Source}
\begin{subfigure}{2.8cm}
\includegraphics[width=2.8cm]{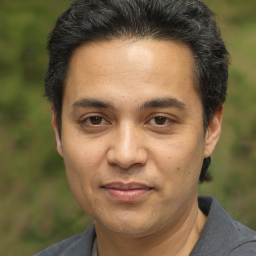}
\end{subfigure}
\begin{subfigure}{2.8cm}
\includegraphics[width=2.8cm]{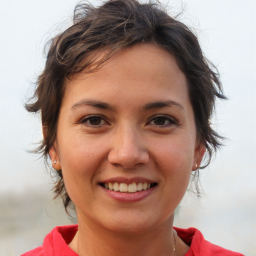}
\end{subfigure}
\begin{subfigure}{2.8cm}
\includegraphics[width=2.8cm]{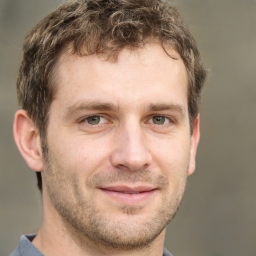}
\end{subfigure}
\begin{subfigure}{2.8cm}
\includegraphics[width=2.8cm]{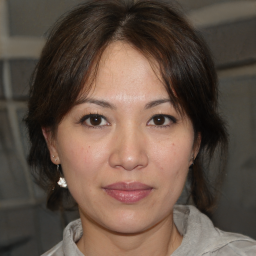}
\end{subfigure}
\begin{subfigure}{2.8cm}
\includegraphics[width=2.8cm]{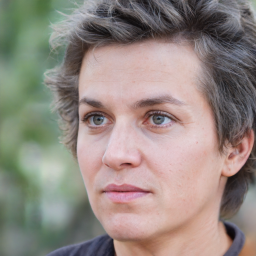}
\end{subfigure}
\begin{subfigure}{2.8cm}
\includegraphics[width=2.8cm]{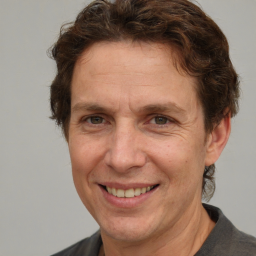}
\end{subfigure}

%\hspace{0.2cm}
%---------row 2
\rotatebox[origin=c]{90}{Reference}
\begin{subfigure}{2.8cm}
\includegraphics[width=2.8cm]{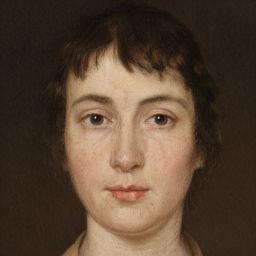}
\end{subfigure}
\begin{subfigure}{2.8cm}
\includegraphics[width=2.8cm]{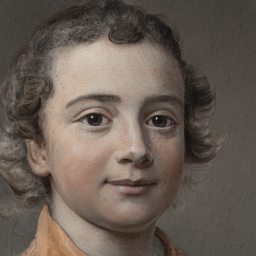}
\end{subfigure}
\begin{subfigure}{2.8cm}
\includegraphics[width=2.8cm]{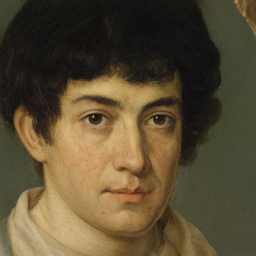}
\end{subfigure}
\begin{subfigure}{2.8cm}
\includegraphics[width=2.8cm]{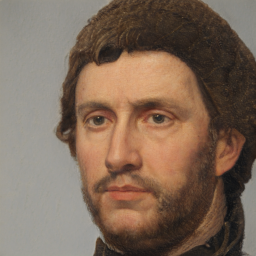}
\end{subfigure}
\begin{subfigure}{2.8cm}
\includegraphics[width=2.8cm]{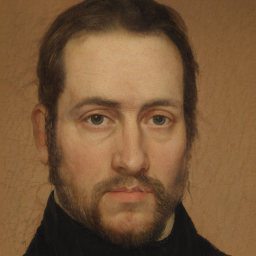}
\end{subfigure}
\begin{subfigure}{2.8cm}
\includegraphics[width=2.8cm]{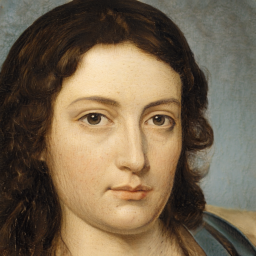}
\end{subfigure}

%---------row 3
\rotatebox[origin=c]{90}{Result}
\begin{subfigure}{2.8cm}
\includegraphics[width=2.8cm]{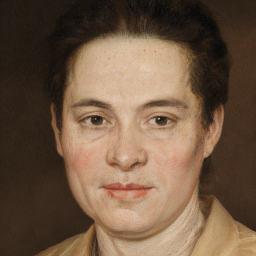}
\end{subfigure}
\begin{subfigure}{2.8cm}
\includegraphics[width=2.8cm]{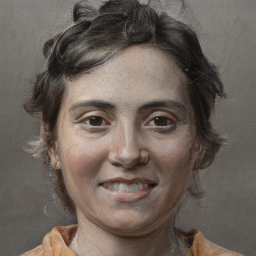}
\end{subfigure}
\begin{subfigure}{2.8cm}
\includegraphics[width=2.8cm]{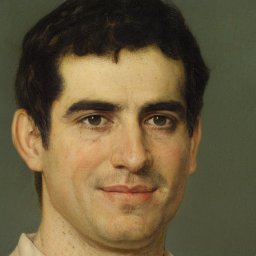}
\end{subfigure}
\begin{subfigure}{2.8cm}
\includegraphics[width=2.8cm]{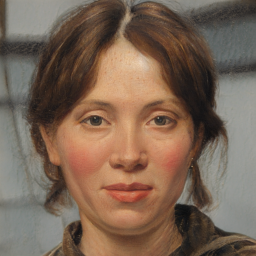}
\end{subfigure}
\begin{subfigure}{2.8cm}
\includegraphics[width=2.8cm]{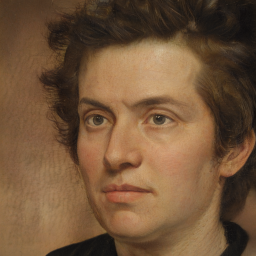}
\end{subfigure}
\begin{subfigure}{2.8cm}
\includegraphics[width=2.8cm]{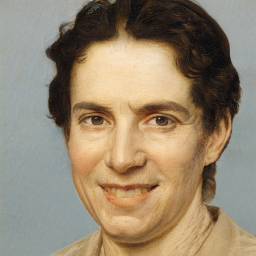}
\end{subfigure}
\caption{Reference guided image generation. The output not only preserves the style of the reference images in terms of color, light condition, texture and so forth, but it also maintains semantic similarity to the source images.}
\label{fig:ref}
\end{figure*}

The proposed method was compared with several state-of-the-art work, and the qualitative results are shown in Fig. \ref{fig:comp_all}. It shows that the proposed method generated images with much better quality, and our results were of higher semantic similarity to the input images. Although the CycleGAN could maintain relatively fine quality, it could generate only a unique image based on an input. Both MUNIT \cite{munit} and DRIT++ \cite{drit++} are unsupervised multi-modal methods. However, the generated was not very close to the input semantically, and we found artifacts in the results. 
% figure comparison
Our method can also perform multi-modal and multi-domain I2I translation, which is shown in Fig. \ref{fig:multi}. A latent vector can be obtained via the proposed inversion method from the image in the source domain. In this case it is the human face domain. We used this latent vector as the content code and randomly sampled five style codes to generate images in anime domain (the first row), portrait domain (the second row) and cartoon domain (the third row) with different styles. Because those models can share the same content code, the translation between every two of those domains can be achieved as well. Fig. \ref{fig:multi} shows that all the images are of high quality and semantic similarity to the source image. In addition, Fig. \ref{fig:ref} shows that the proposed method can also generate images based on a reference image provided by a user. In that case, the latent code of the source image is used as the content code, and the inverse latent code from the reference is then adopted as the style code for generation. The generated images then inherit the style from the reference image, and they also maintain the semantic structure of the source image.\par
% figure multi-modal
% figure reference guided
\subsection{Ablation study and Analysis}
\begin{figure}[!ht]
   \centering
%---------row 1
\begin{subfigure}{2.8cm}
\includegraphics[width=2.8cm]{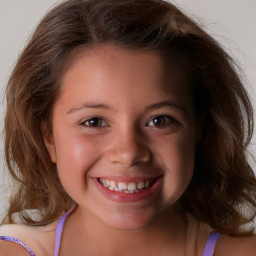}
\end{subfigure}
\begin{subfigure}{2.8cm}
\includegraphics[width=2.8cm]{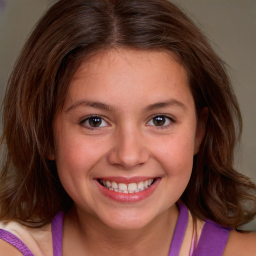}
\end{subfigure}
\begin{subfigure}{2.8cm}
\includegraphics[width=2.8cm]{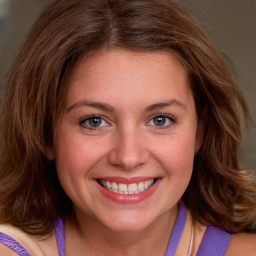}
\end{subfigure}

%---------row 1
\begin{subfigure}{2.8cm}
\includegraphics[width=2.8cm]{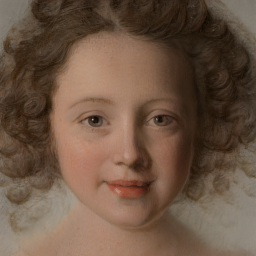}\caption*{\small{Age-}}
\end{subfigure}
\begin{subfigure}{2.8cm}
\includegraphics[width=2.8cm]{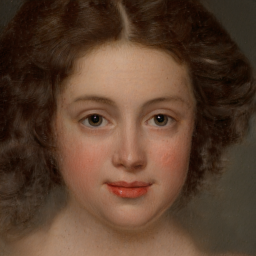}\caption*{\small{Age}}
\end{subfigure}
\begin{subfigure}{2.8cm}
\includegraphics[width=2.8cm]{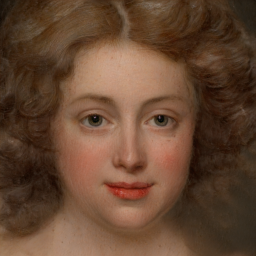}\caption*{\small{
Age+}}
\end{subfigure}
\caption{GAN space invariance. After transformation, the GAN space of the base model can be inherited by the generated model. We found the age direction in the base model. By moving along this direction, the generated model (portrait model in the second row) can generate images at different ages as well.}
\label{fig:ganspace}
\end{figure}

\begin{figure*}[!ht]
   \centering
%---------row 1
\begin{subfigure}{2.8cm}
\includegraphics[width=2.8cm]{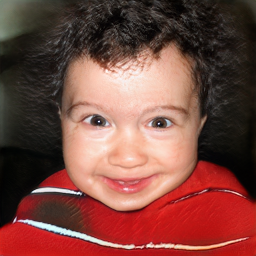}
\end{subfigure}
\begin{subfigure}{2.8cm}
\includegraphics[width=2.8cm]{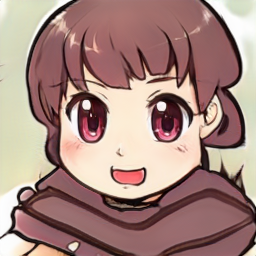}
\end{subfigure}
\begin{subfigure}{2.8cm}
\includegraphics[width=2.8cm]{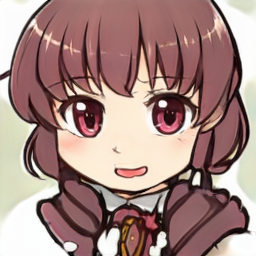}
\end{subfigure}
\begin{subfigure}{2.8cm}
\includegraphics[width=2.8cm]{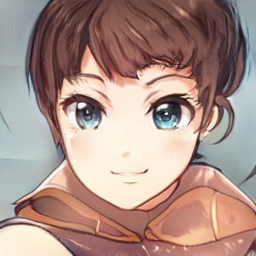}
\end{subfigure}
\begin{subfigure}{2.8cm}
\includegraphics[width=2.8cm]{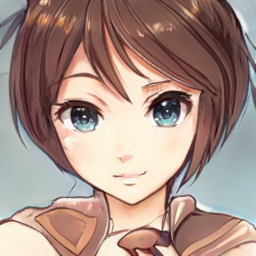}
\end{subfigure}

%---------row 2
\begin{subfigure}{2.8cm}
\includegraphics[width=2.8cm]{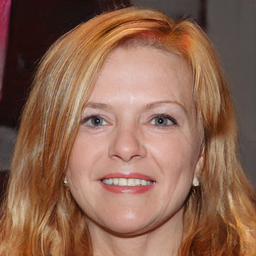}
\end{subfigure}
\begin{subfigure}{2.8cm}
\includegraphics[width=2.8cm]{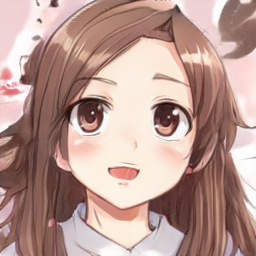}
\end{subfigure}
\begin{subfigure}{2.8cm}
\includegraphics[width=2.8cm]{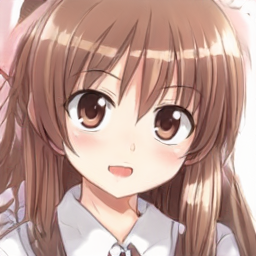}
\end{subfigure}
\begin{subfigure}{2.8cm}
\includegraphics[width=2.8cm]{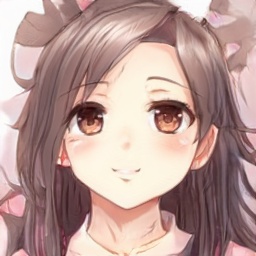}
\end{subfigure}
\begin{subfigure}{2.8cm}
\includegraphics[width=2.8cm]{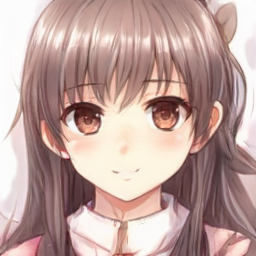}
\end{subfigure}

%---------row 3
\begin{subfigure}{2.8cm}
\includegraphics[width=2.8cm]{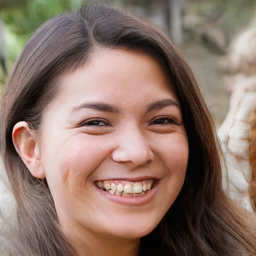}
\end{subfigure}
\begin{subfigure}{2.8cm}
\includegraphics[width=2.8cm]{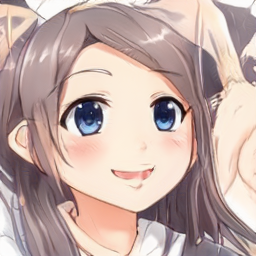}
\end{subfigure}
\begin{subfigure}{2.8cm}
\includegraphics[width=2.8cm]{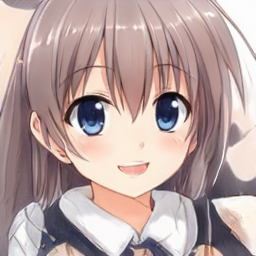}
\end{subfigure}
\begin{subfigure}{2.8cm}
\includegraphics[width=2.8cm]{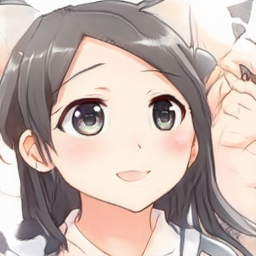}
\end{subfigure}
\begin{subfigure}{2.8cm}
\includegraphics[width=2.8cm]{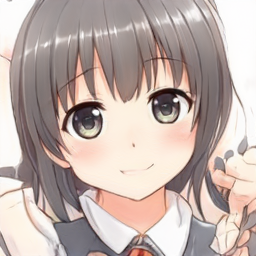}
\end{subfigure}

%---------row 4
\begin{subfigure}{2.8cm}
\includegraphics[width=2.8cm]{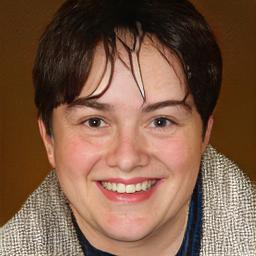}\caption*{\small{Source domain}}
\end{subfigure}
\begin{subfigure}{2.8cm}
\includegraphics[width=2.8cm]{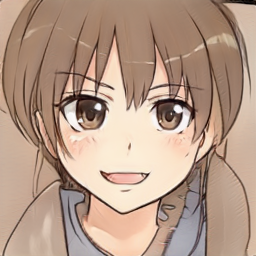}\caption*{\small{FC-FT-LS}}
\end{subfigure}
\begin{subfigure}{2.8cm}
\includegraphics[width=2.8cm]{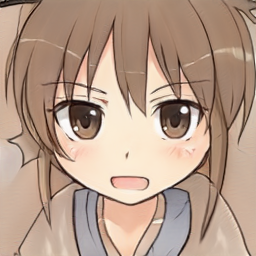}\caption*{\small{FC-FT}}
\end{subfigure}
\begin{subfigure}{2.8cm}
\includegraphics[width=2.8cm]{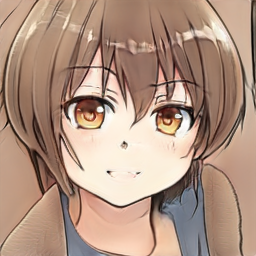}\caption*{\small{FT-LS}}
\end{subfigure}
\begin{subfigure}{2.8cm}
\includegraphics[width=2.8cm]{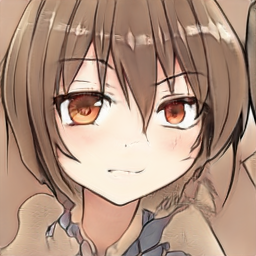}\caption*{\small{FT}}
\end{subfigure}
\caption{Ablation study of different transformation set-ups. The combination of FC-FT-LS achieves the best results (the second column). It presents the highest semantic similarity to the source images including features like age, expression, hair and shape of the face, compared to other results.}
\label{fig:abl_fcls}
\end{figure*}

\textbf{Ablation Study of Fine-tuning} First, we found that model fine-tuning gave the fine-tuned model a GAN space that was similar to the base model. Thus, we used the semantic space introduced by \cite{closedform} and found the directions controlling ages in the base model (styleGAN2 model trained on FFHQ \cite{ffhq}). Fig. \ref{fig:ganspace} shows that, the fine-tuned model presents similar changes in terms of age at those directions.Moreover, we conducted an ablation study of the proposed freeze-FC method. (We refer to the fine-tuning process with freeze-FC as FC, fine-tune as FT and LS as layer-swap.) In Fig. \ref{fig:abl_fcls}, generated images with FC could have more similarities to the source domain. Freezing the FC layers  preserves the mapping relation from a latent code to an input feature vector. Therefore, some semantic information like age and facial expression can be preserved. For example, in the first row of Fig. \ref{fig:abl_fcls}, the generated images from FC-FT are more like a child than the results from FT. Similarly, results from FC-FT/FC-FT-LS match the facial expression of the source domain with an open mouth while the results from FT can  maintain only the rough pose. \par
% figure ganspace
% figure fc-ft-ls ablation
\begin{figure*}[!ht]
   \centering
%---------row 1
\begin{subfigure}{2.8cm}
\includegraphics[width=2.8cm]{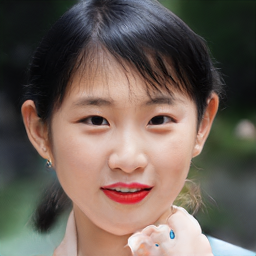}
\end{subfigure}
\begin{subfigure}{2.8cm}
\includegraphics[width=2.8cm]{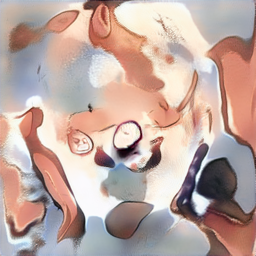}
\end{subfigure}
\begin{subfigure}{2.8cm}
\includegraphics[width=2.8cm]{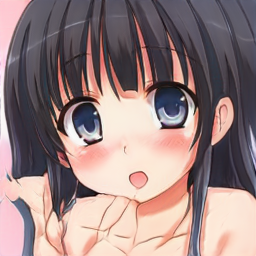}
\end{subfigure}
\begin{subfigure}{2.8cm}
\includegraphics[width=2.8cm]{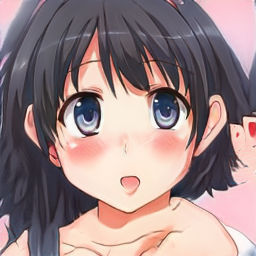}
\end{subfigure}
\begin{subfigure}{2.8cm}
\includegraphics[width=2.8cm]{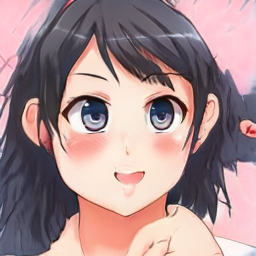}
\end{subfigure}
\begin{subfigure}{2.8cm}
\includegraphics[width=2.8cm]{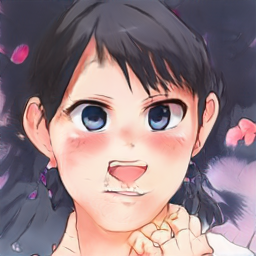}
\end{subfigure}

%---------row 2
\begin{subfigure}{2.8cm}
\includegraphics[width=2.8cm]{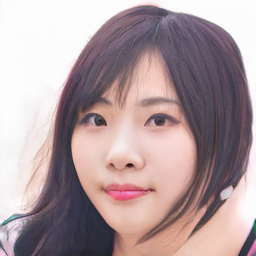}\vspace{0.1cm}
\end{subfigure}
\begin{subfigure}{2.8cm}
\includegraphics[width=2.8cm]{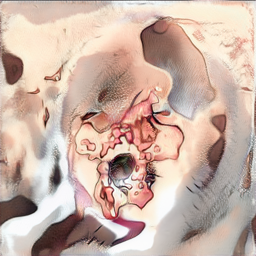}\vspace{0.1cm}
\end{subfigure}
\begin{subfigure}{2.8cm}
\includegraphics[width=2.8cm]{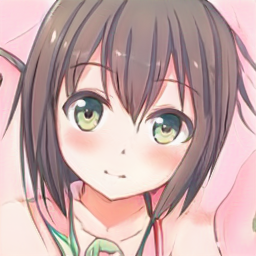}\vspace{0.1cm}
\end{subfigure}
\begin{subfigure}{2.8cm}
\includegraphics[width=2.8cm]{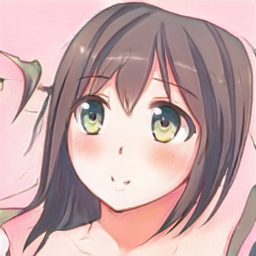}\vspace{0.1cm}
\end{subfigure}
\begin{subfigure}{2.8cm}
\includegraphics[width=2.8cm]{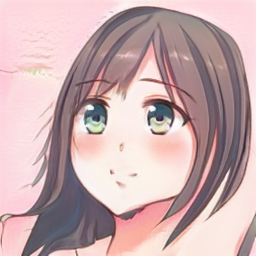}\vspace{0.1cm}
\end{subfigure}
\begin{subfigure}{2.8cm}
\includegraphics[width=2.8cm]{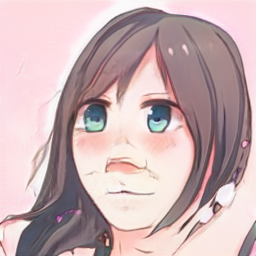}\vspace{0.1cm}
\end{subfigure}

%---------row 3
\begin{subfigure}{2.8cm}
\includegraphics[width=2.8cm]{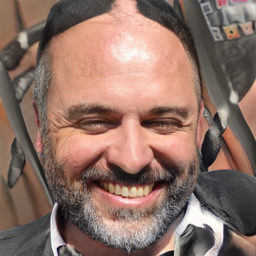}
\end{subfigure}
\begin{subfigure}{2.8cm}
\includegraphics[width=2.8cm]{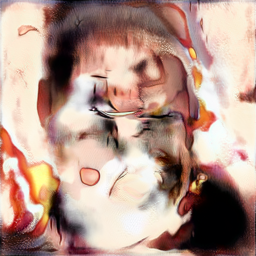}
\end{subfigure}
\begin{subfigure}{2.8cm}
\includegraphics[width=2.8cm]{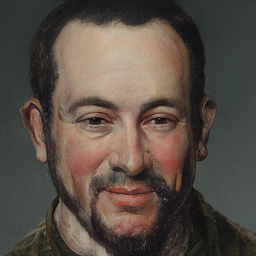}
\end{subfigure}
\begin{subfigure}{2.8cm}
\includegraphics[width=2.8cm]{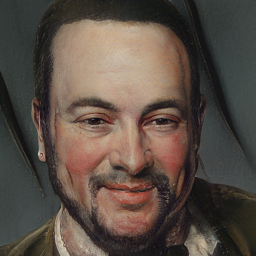}
\end{subfigure}
\begin{subfigure}{2.8cm}
\includegraphics[width=2.8cm]{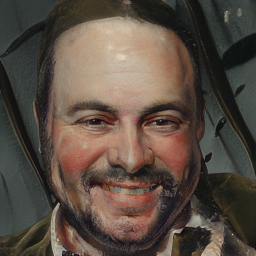}
\end{subfigure}
\begin{subfigure}{2.8cm}
\includegraphics[width=2.8cm]{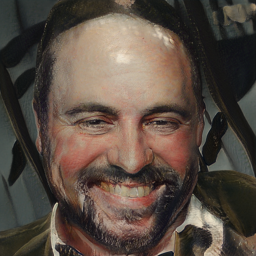}
\end{subfigure}

%---------row 4
\begin{subfigure}{2.8cm}
\includegraphics[width=2.8cm]{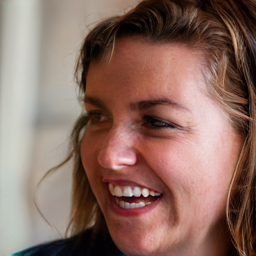}\vspace{0.1cm}
\end{subfigure}
\begin{subfigure}{2.8cm}
\includegraphics[width=2.8cm]{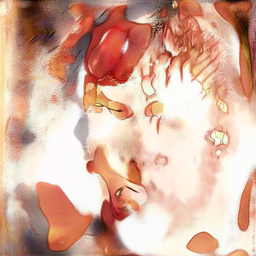}\vspace{0.1cm}
\end{subfigure}
\begin{subfigure}{2.8cm}
\includegraphics[width=2.8cm]{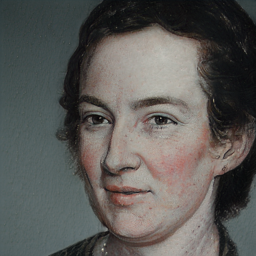}\vspace{0.1cm}
\end{subfigure}
\begin{subfigure}{2.8cm}
\includegraphics[width=2.8cm]{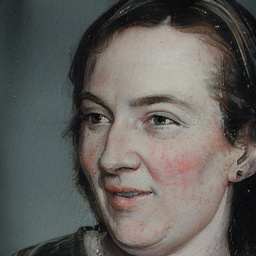}\vspace{0.1cm}
\end{subfigure}
\begin{subfigure}{2.8cm}
\includegraphics[width=2.8cm]{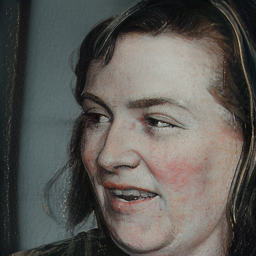}\vspace{0.1cm}
\end{subfigure}
\begin{subfigure}{2.8cm}
\includegraphics[width=2.8cm]{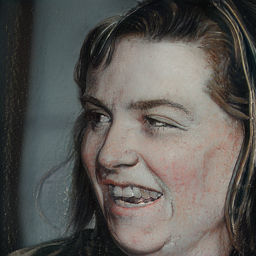}\vspace{0.1cm}
\end{subfigure}

%---------row 5
\begin{subfigure}{2.8cm}
\includegraphics[width=2.8cm]{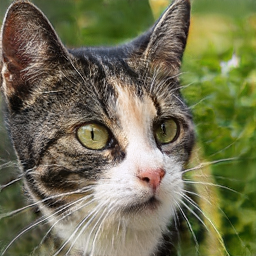}
\end{subfigure}
\begin{subfigure}{2.8cm}
\includegraphics[width=2.8cm]{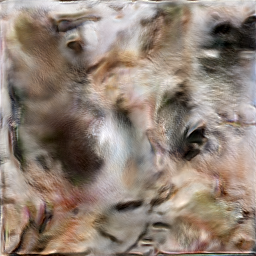}
\end{subfigure}
\begin{subfigure}{2.8cm}
\includegraphics[width=2.8cm]{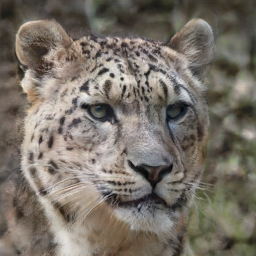}
\end{subfigure}
\begin{subfigure}{2.8cm}
\includegraphics[width=2.8cm]{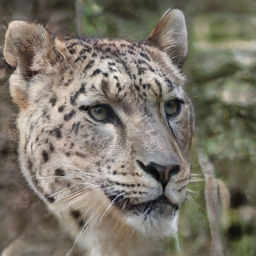}
\end{subfigure}
\begin{subfigure}{2.8cm}
\includegraphics[width=2.8cm]{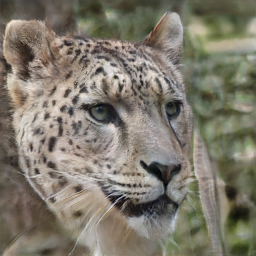}
\end{subfigure}
\begin{subfigure}{2.8cm}
\includegraphics[width=2.8cm]{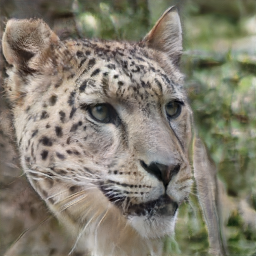}
\end{subfigure}

%---------row 6
\begin{subfigure}{2.8cm}
\includegraphics[width=2.8cm]{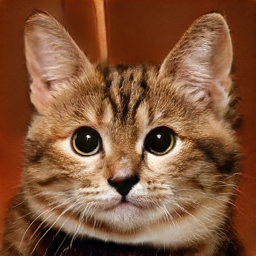}\caption*{\small{Source domain}}
\end{subfigure}
\begin{subfigure}{2.8cm}
\includegraphics[width=2.8cm]{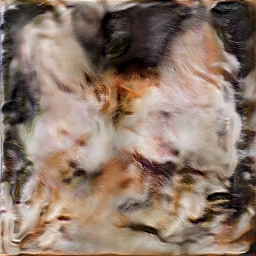}\caption*{\small{Direct-LS}}
\end{subfigure}
\begin{subfigure}{2.8cm}
\includegraphics[width=2.8cm]{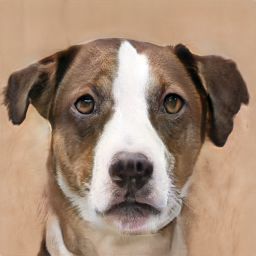}\caption*{\small{LS=0}}
\end{subfigure}
\begin{subfigure}{2.8cm}
\includegraphics[width=2.8cm]{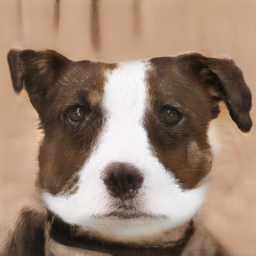}\caption*{\small{LS=1}}
\end{subfigure}
\begin{subfigure}{2.8cm}
\includegraphics[width=2.8cm]{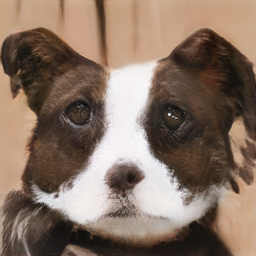}\caption*{\small{LS=3}}
\end{subfigure}
\begin{subfigure}{2.8cm}
\includegraphics[width=2.8cm]{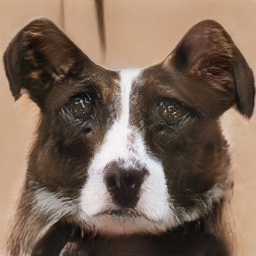}\caption*{\small{LS=5}}
\end{subfigure}
\caption{Layer-swap analysis. We found direct layer-swap between two models without model transformation would not work (the second column). With more layers used in layer-swap, the generated output is more like the source image. However, as layer-swap tries to add features in one domain to images in another domain, the generated output might be corrupted since they are beyond the target domain when there is a large distance between the two domains such as a small intersection space. This is proved in face2anime (first two rows) and cat2dog (last two rows). When $LS=5$, the output seems unnatural as it has many artifacts. }
\label{fig:swap_ana}
\end{figure*}

\textbf{Ablation Study of Layer-swap} Then we analyzed the results with different levels of layer-swap. This is shown in Fig. \ref{fig:swap_ana}, where Direct-LS represents layer-swap without model fine-tuning, while $LS=n$ presents the number of swapped layers. $LS=0$ indicates model transformation without layer-swap. Fig. \ref{fig:swap_ana} shows that all the images generated with Direct-LS are corrupted. That is because that feature cannot be shared between two models with relatively large model distance. As mentioned before, fine-tuning could reduce the model distance, and those high-level features can then be re-used by similar models. The results also indicate that with more layer-swapping, the generated images could be more like the source domain images. However, this property cannot hold if there is a large distance between the domains themselves. For example, the anime domain is quite different from the face domain in terms of the shape of the face and mouth, and the anime does not have obvious human-like features such as noses. In such cases, with a large amount of layer-swapping, the generated output would contain artifacts (the first two rows, $LS=5$), and they would even be out of the target domain (the last row, $LS=5$). \par
% figure layer-swap
\textbf{Inversion Comparison.} We also compared the proposed inversion method with several state-of-the-art work including the projection method proposed by styleGAN2 \cite{stylegan2}, In-domain invert encoding and inversion method \cite{indomain} and Image2StyleGAN \cite{image2stylegan}. The results are listed in Table. \ref{tbl:inv}, which shows that the proposed method has better reconstruction quality compared to In-domain invert methods and the Image2StyleGAN \cite{image2stylegan}. The proposed method has some superiority compared to the styleGAN2 \cite{stylegan2} projection method. This is proved in the quantitative results.

\begin{figure*}[!ht]
   \centering
%---------row 1
\begin{subfigure}{2.8cm}
\includegraphics[width=2.8cm]{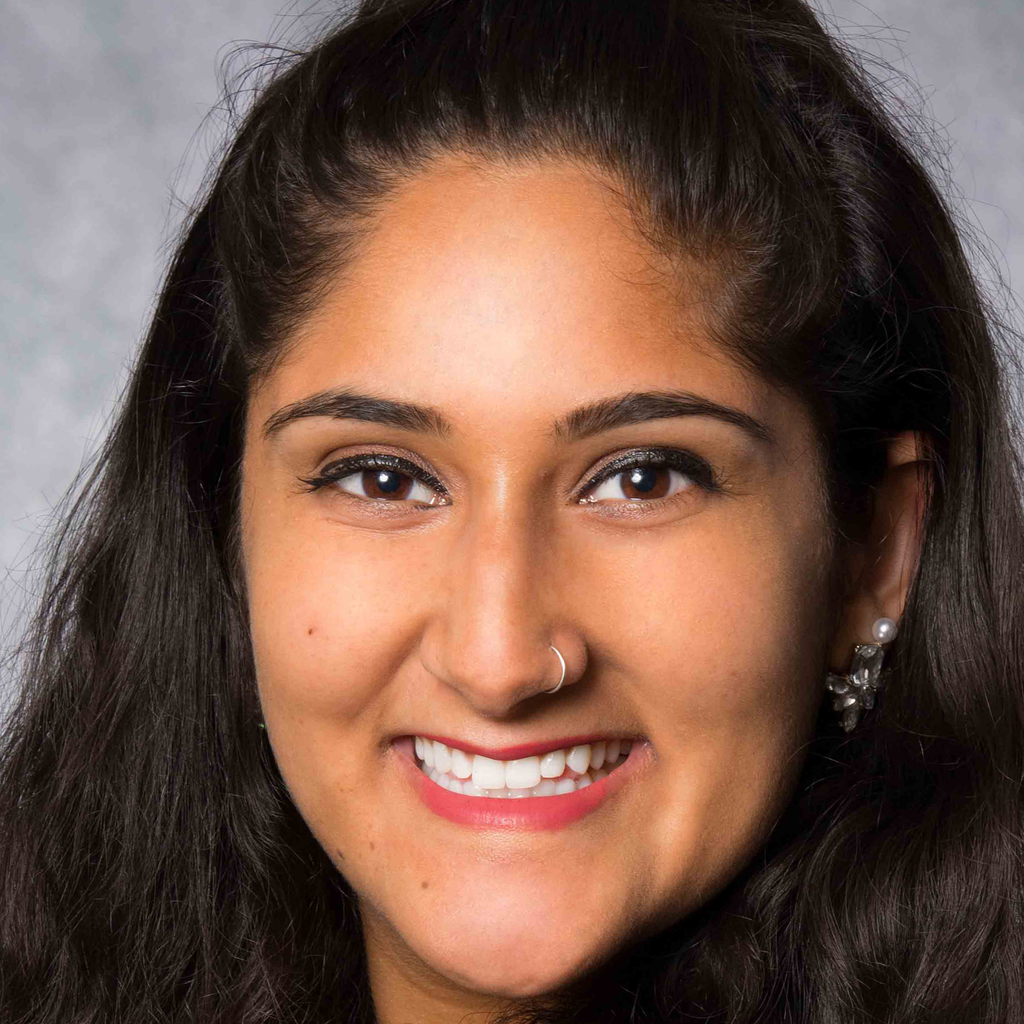}
\end{subfigure}
\begin{subfigure}{2.8cm}
\includegraphics[width=2.8cm]{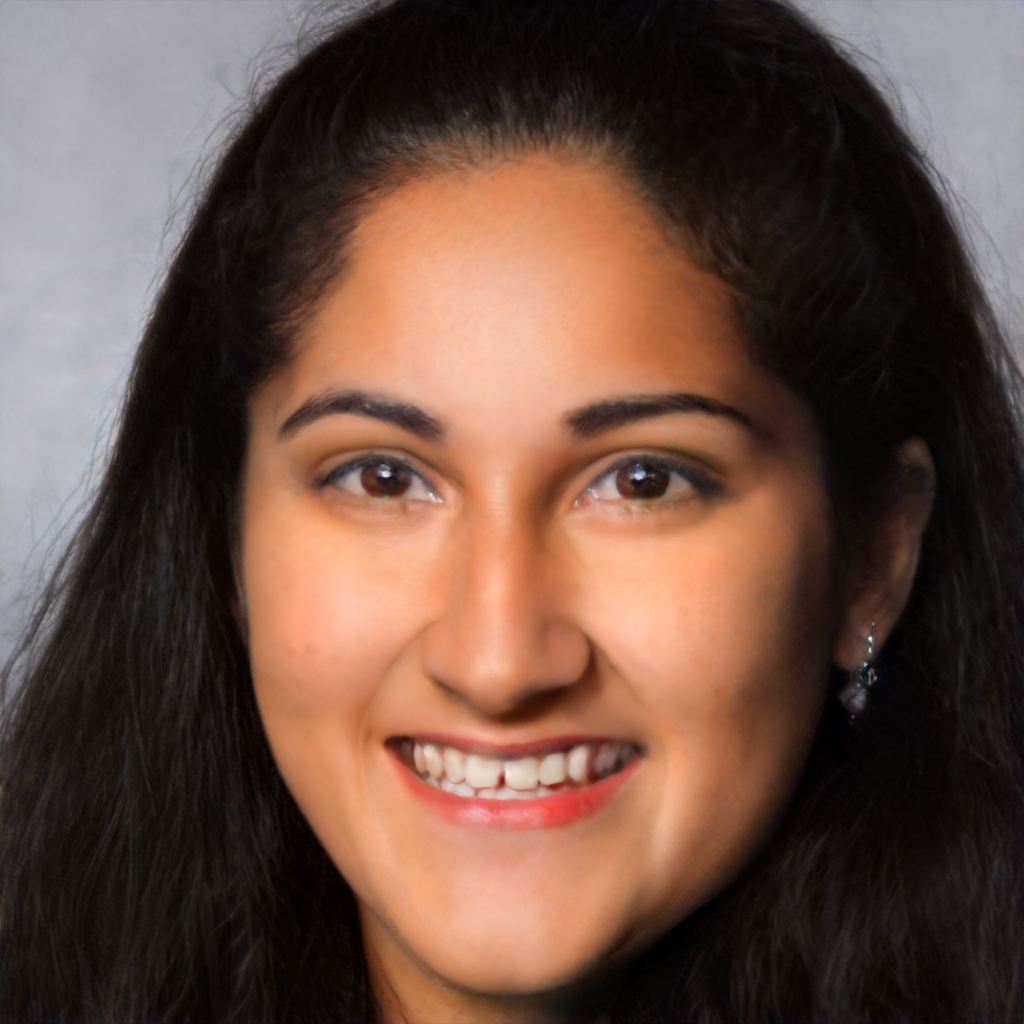}
\end{subfigure}
\begin{subfigure}{2.8cm}
\includegraphics[width=2.8cm]{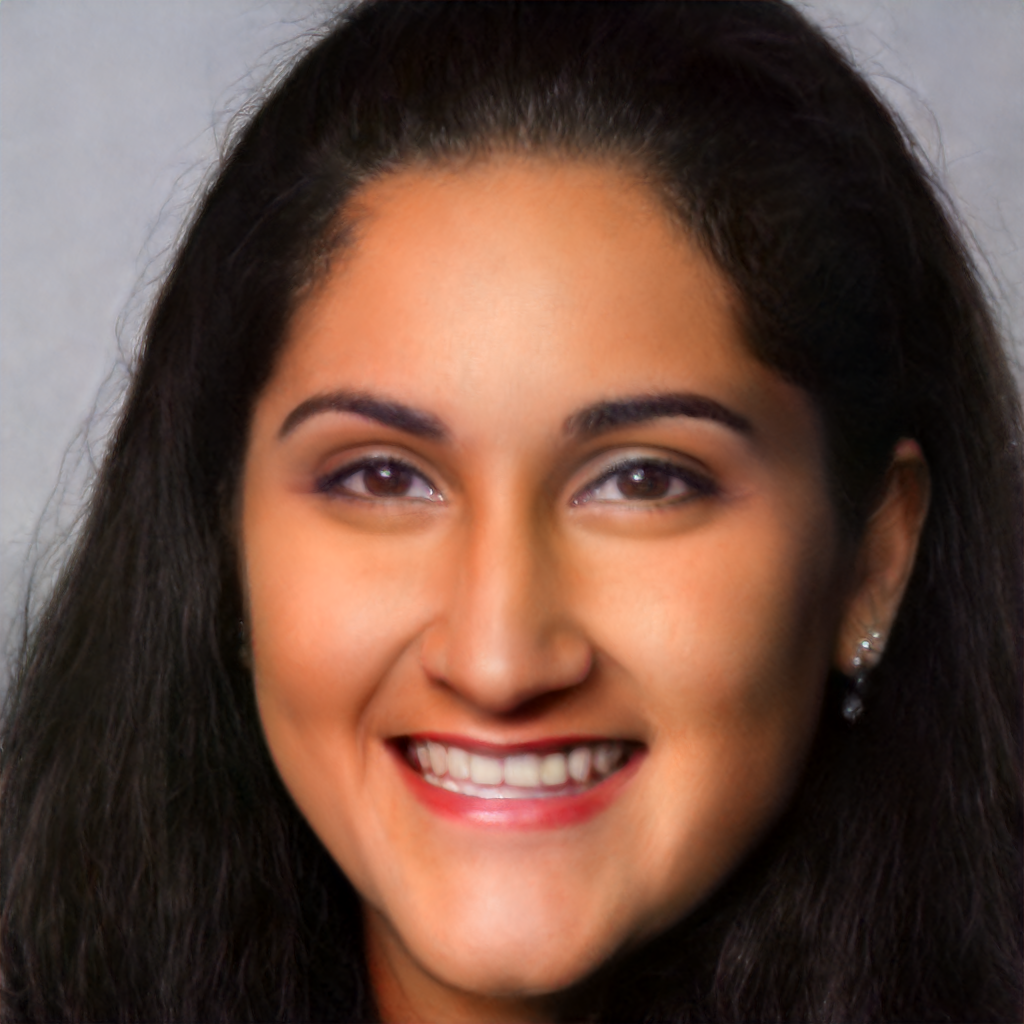}
\end{subfigure}
\begin{subfigure}{2.8cm}
\includegraphics[width=2.8cm]{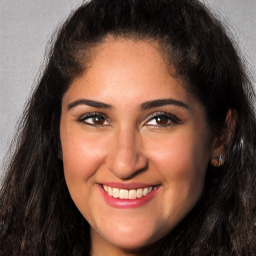}
\end{subfigure}
\begin{subfigure}{2.8cm}
\includegraphics[width=2.8cm]{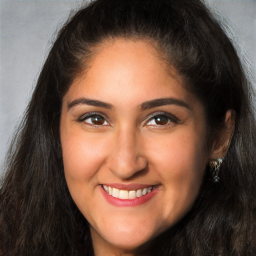}
\end{subfigure}
\begin{subfigure}{2.8cm}
\includegraphics[width=2.8cm]{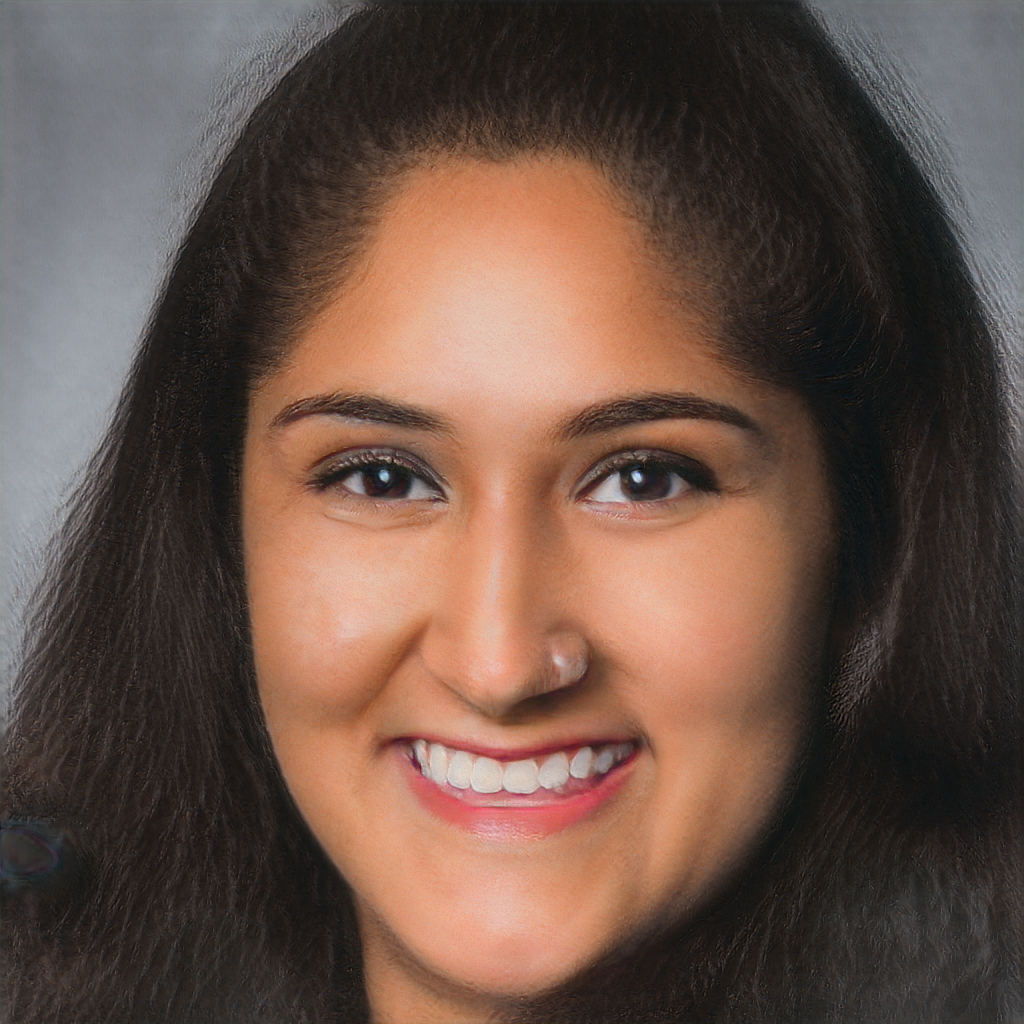}
\end{subfigure}

%---------row 2 
\begin{subfigure}{2.8cm}
\includegraphics[width=2.8cm]{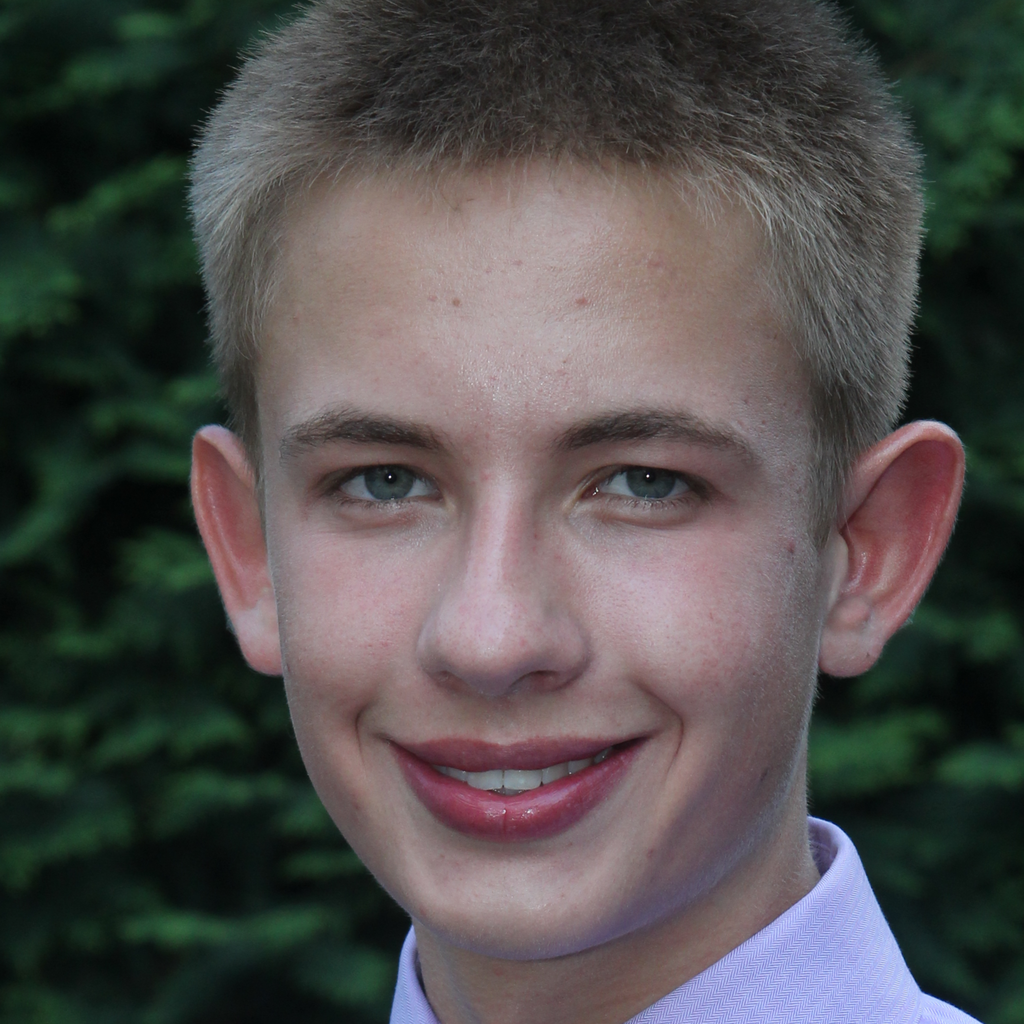}\caption*{\small{Ground Truth}}
\end{subfigure}
\begin{subfigure}{2.8cm}
\includegraphics[width=2.8cm]{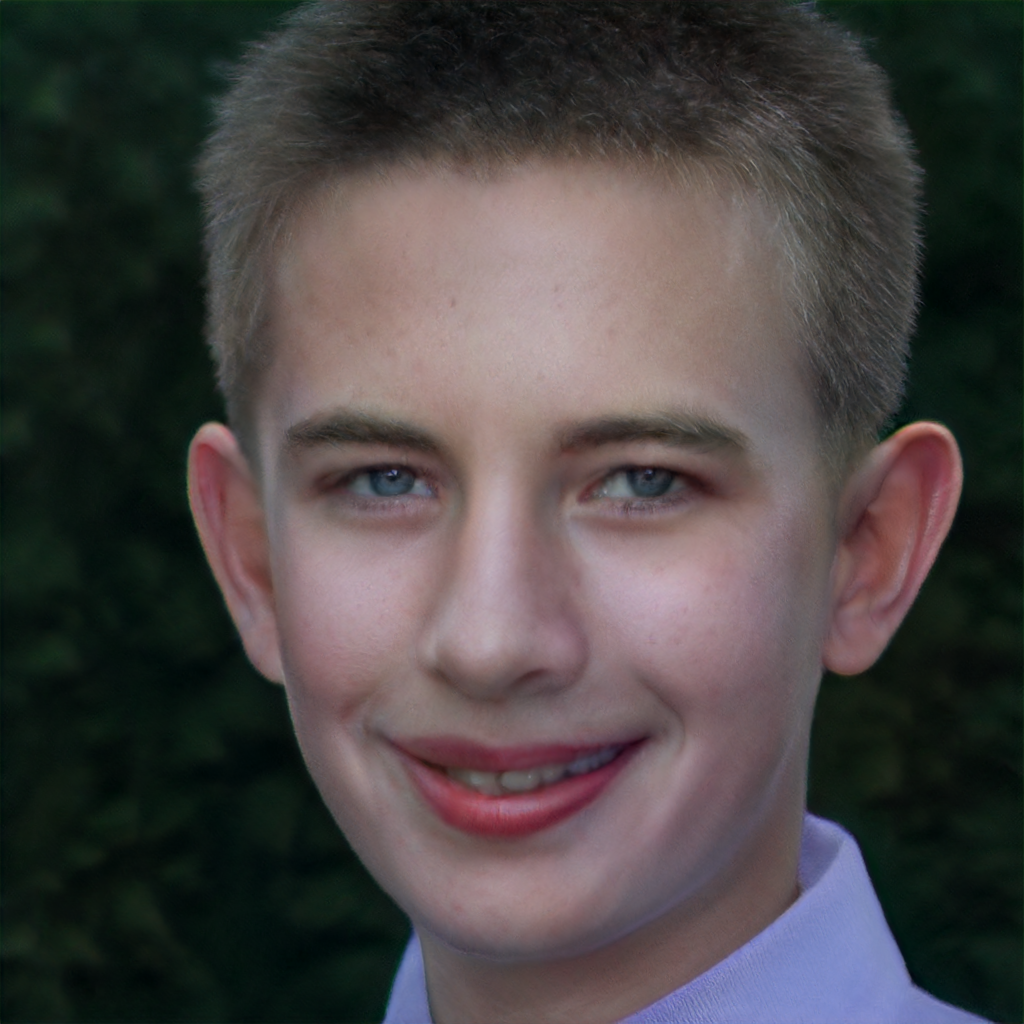}\caption*{\small{Ours}}
\end{subfigure}
\begin{subfigure}{2.8cm}
\includegraphics[width=2.8cm]{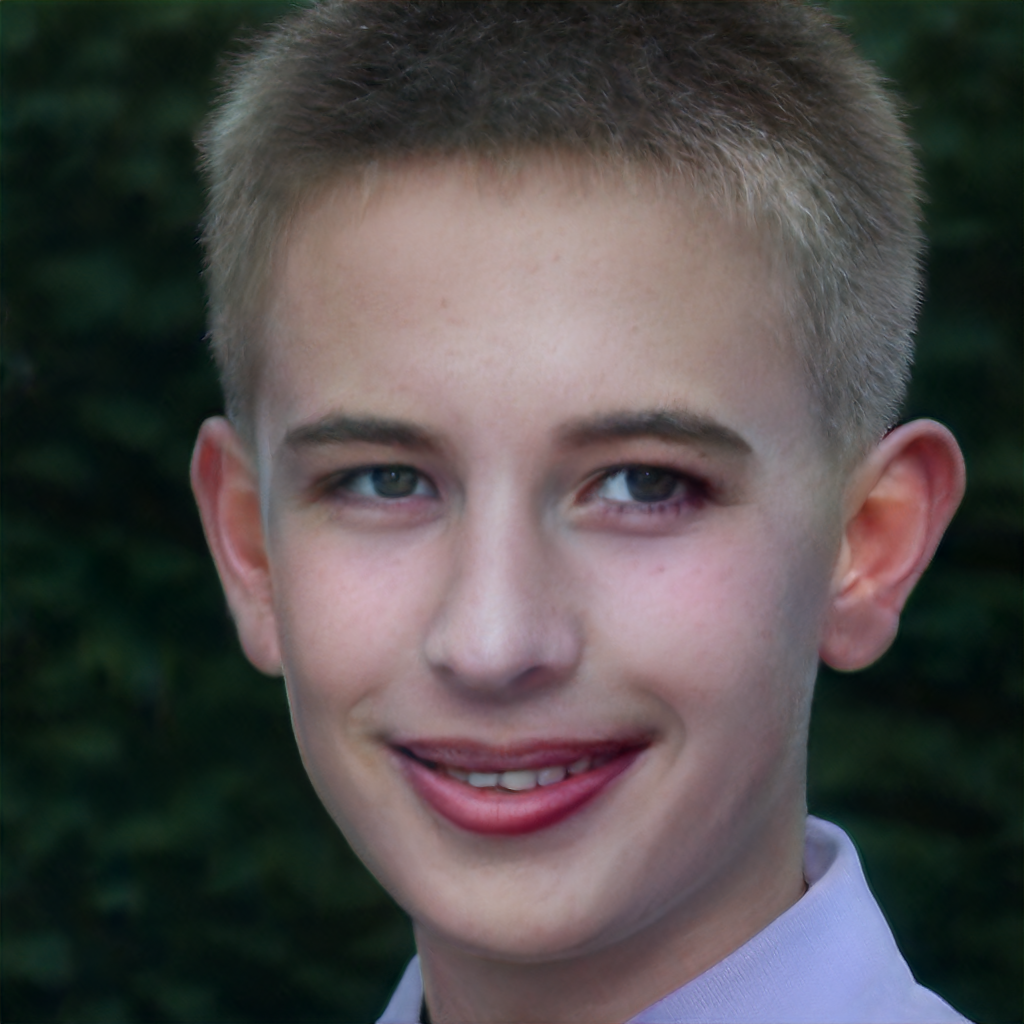}\caption*{\small{StyleGAN2-Proj}}
\end{subfigure}
\begin{subfigure}{2.8cm}
\includegraphics[width=2.8cm]{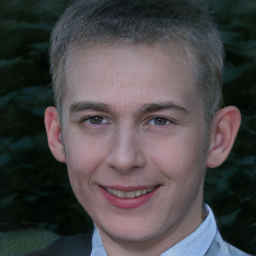}\caption*{\small{IdInvert-Enc}}
\end{subfigure}
\begin{subfigure}{2.8cm}
\includegraphics[width=2.8cm]{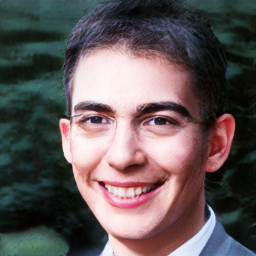}\caption*{\small{IdInvert-Inv}}
\end{subfigure}
\begin{subfigure}{2.8cm}
\includegraphics[width=2.8cm]{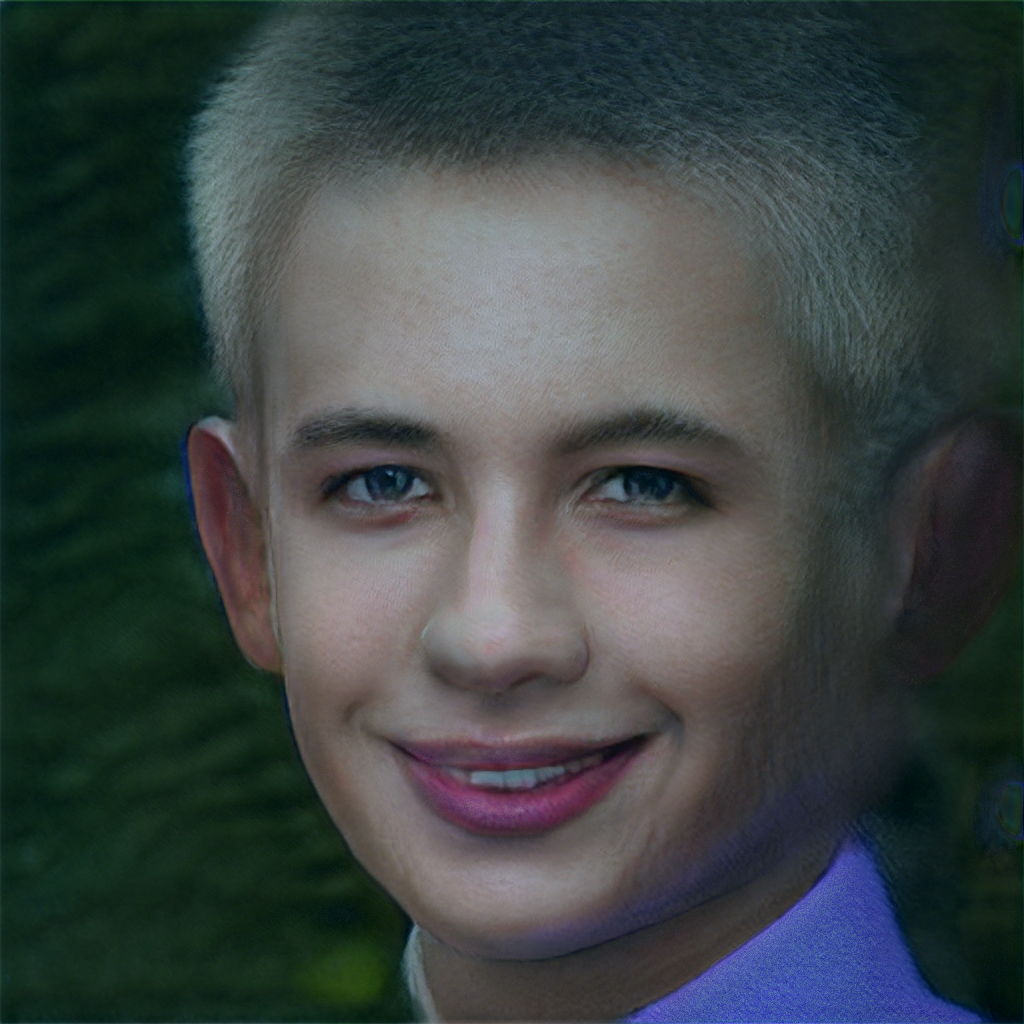}\caption*{\small{Image2StyleGAN}}
\end{subfigure}

\caption{Comparison of inversion methods. The proposed method is compared with other state-of-the-art inversion methods. It is obvious that our method and the StyleGAN2 projection method achieve better reconstruction results than the others, while our method also presents a slightly greater similarity to the ground truth image than the StyleGAN2 projection method. This is also proved by SSIM score in Table \ref{tbl:inv}}
\label{fig:inv_comp}
\end{figure*}

\subsection{Quantitative Results}
The model distance was defined in Eq. \ref{eqn:dist}, specifically we used LPIPS \cite{lpips} to measure the distance between two images generated by two models. To perform the quantitative evaluation, we sampled 2000 latent vector $z$ and fed them into the compared models to generate corresponding output. Then we computed the average LPIPS scores of the 2000 pairs of images. We found the proposed model transformation could drastically reduce the model distances as shown in Table. \ref{tbl:model_distance}. It is shown that the model distance is relatively large when they are trained independently (the first row), and FC-FT can achieve closer distance compared to direct FT. Moreover, layer-swap can also diminish the distance, and the more layers that are swapped, the smaller model distance will be. Furthermore, the model distances shown in Table \ref{tbl:model_distance} indicate that the original domain gap between the different dataset are different. For example, the distance between face and anime is still larger than that between face and portrait. This matches our perceptual understanding that portrait looks more like the images of face compared to anime. Besides, we found that the distance between the cartoon model and the portrait model transformed from the face model was around 0.434. Though this is a little bit higher than the results of direct transformation (0.398), it is much smaller than the original distance (0.692). This proves that the proposed method can support multi-domain I2I translation.\par
% model distance table
\begin{table*}[!ht]
\setlength{\tabcolsep}{3.5mm}{
\begin{tabular}{ccccccc}
\hline
 & \textbf{face2anime} & \textbf{face2cartoon} & \textbf{face2portrait} & \textbf{cartoon2metface} & \textbf{cat2dog} & \textbf{cat2wild}  \\ \hline
w/o FT &  0.747& 0.632 &0.759  & 0.692 & 0.672 & 0.593  \\ \hline
FT & 0.650 & 0.551 & 0.636 & 0.627 & 0.625 & 0.570  \\ \hline
FC-FT & 0.639 & 0.483 & 0.612 & 0.587 & 0.599 & 0.568  \\ \hline
\begin{tabular}[c]{@{}l@{}}FC-FT-LS1\end{tabular} & 0.610 & 0.450 & 0.515 & 0.537 & 0.598 & 0.567  \\ \hline
\begin{tabular}[c]{@{}l@{}}FC-FT-LS3\end{tabular} &  0.571& 0.398 & 0.455 & 0.478 &0.595  & 0.561 \\ \hline
\begin{tabular}[c]{@{}l@{}}FC-FT-LS5\end{tabular} &\textbf{0.548} & \textbf{0.315} & \textbf{0.379} & \textbf{0.398} &\textbf{0.581}  &\textbf{0.557}   \\ \hline
\caption{Model Distance between two models on different datasets. We refer to the fine-tuning process with freeze-FC as FC, fine-tune as FT and LS as layer-swap, where the number following LS indicates the number of swapped layers.}
\label{tbl:model_distance}
\end{tabular}
}
\end{table*}
We performed quantitative evaluation on the proposed method and the base line models. The results are shown in Table \ref{tbl:scores}. FID \cite{fid} was employed to evaluate the quality of the generated results, where $2\,000$ test images were randomly chosen as the input. We calculated the average FID between the output and images in the target dataset. To evaluate the diversity of generated images, we computed the LPIPS distance between two randomly selected images in the generated results, which is referred as $LPIPS_d$. And we computed LPIPS between the generated image and the input to assess the semantic similarity, which is represented by $LPIPS_s$. The proposed method, including FC-FT and FC-FT-LS, can achieve the best scores on both portrait and anime datasets. For FID and diversity, the proposed method without layer-swap could obtain a slightly better score. That is because the generated image might not be as similar as the results without swapping to the images in the target domain. As for $LPIPS_s$, our method with layer-swap achieved the best scores.\par
% comparison scores table
\begin{table*}[!ht]
\begin{tabular}{lllllll}
\hline
\multirow{2}{*}{} & \multicolumn{3}{l}{\textbf{face2portrait}} & \multicolumn{3}{l}{\textbf{face2anime}} \\ \cline{2-7} 
 & $FID\downarrow$ & $LPIPS_d\uparrow$ & $LPIPS_s\downarrow$ & $FID\downarrow$ & $LPIPS_d\uparrow$ & $LPIPS_s\downarrow$ \\ \hline
CycleGAN\cite{cyclegan} &  54.472& 0.649 & 0.587 &  78.452& 0.645 & 0.598 \\ \hline
MUNIT\cite{munit} & 62.683 & 0.650 & 0.677 &  66.261& 0.749 & 0.770 \\ \hline
DRIT++\cite{drit++} &  76.271& 0.706 & 0.667 &  72.293& 0.647  & 0.751 \\ \hline
Ours (FC-FT) &  \textbf{32.480}& \textbf{0.743} & 0.489 & \textbf{52.930} & \textbf{0.767} & 0.591 \\ \hline
Ours (FC-FT-LS) &34.147  & 0.742 & \textbf{0.468} &  53.737&\textbf{0.767}  &\textbf{0.573}  \\ \hline
\caption{FID \cite{fid}  and LPIPS \cite{lpips} scores for the proposed method and other state-of-the-art work. We refer to the fine-tuning process with freeze-FC as FC, fine-tune as FT and LS as layer-swap. We choose $swap-layer=3$ for anime data and $swap-layer=5$ for portrait data as the anime data has relatively larger distance to the face data.}
\label{tbl:scores}
\end{tabular}
\end{table*}

We also compared the proposed method with several state-of-the-art work quantitatively, where SSIM \cite{ssim} was adopted as the metric to evaluate reconstruction accuracy. The results in Table \ref{tbl:inv} show that our method obtains a much higher score than In-domain inversion and Image2StyleGAN \cite{image2stylegan}. Compared to the projection method (0.655), the proposed inversion method also achieved a slightly higher score (0.667).\par
% SSIM score table
\begin{table}[]
\setlength{\tabcolsep}{7mm}{
\begin{tabular}{cc}
\hline
\textbf{Method} & \textbf{SSIM $\uparrow$} \\ \hline
Image2StyleGAN \cite{image2stylegan} & 0.613 \\ \hline
IdInvert-inv \cite{indomain} & 0.623 \\ \hline
IdInvert-enc \cite{indomain} & 0.587 \\ \hline
StyleGan2-proj \cite{stylegan2} & 0.655 \\ \hline
\textbf{Ours} & \textbf{0.667} \\ \hline
\caption{SSIM \cite{ssim} scores for different inversion methods}
\label{tbl:inv}
\end{tabular}
}
\end{table}